\documentclass[preprint,times,review]{elsarticle}

\usepackage{amsmath,amssymb,amsthm,upgreek}
\usepackage{epsfig}
\usepackage{graphicx}
\usepackage{makecell}
\usepackage{subfigure}
\usepackage{times}
\usepackage{epstopdf}
\usepackage{booktabs}
\usepackage{caption}

\usepackage{url}
\usepackage{hyperref}
\usepackage{float}

\usepackage{verbatim}
\usepackage{algorithm}
\usepackage{algorithmic}
\usepackage{bm}

\usepackage{threeparttable}
\usepackage{multicol}
\usepackage{multirow}
\usepackage{enumerate}

\usepackage{color}

\usepackage{lineno}
\journal{Pattern Recognition}

\begin{document}
	\captionsetup[figure]{name={Fig.}}
	\begin{frontmatter}

		\title{Unsupervised Domain Adaptation via Distilled Discriminative Clustering}

        \author[scut]{Hui~Tang}
        \ead{eehuitang@mail.scut.edu.cn}
        \author[pclab]{Yaowei~Wang}
        \ead{yaoweiwang@pku.edu.cn}
        \author[scut]{Kui~Jia\corref{cor1}}
        \ead{kuijia@scut.edu.cn}
		
		\cortext[cor1]{Corresponding author.}
        \address[scut]{School of Electronic and Information Engineering, South China University of Technology, \\ Guangzhou 510641, Guangdong, China}
        \address[pclab]{Peng Cheng Laboratory, Shenzhen 518000, Guangdong, China}
		
		\begin{abstract}
		    Unsupervised domain adaptation addresses the problem of classifying data in an unlabeled target domain, given labeled source domain data that share a common label space but follow a different distribution. Most of the recent methods take the approach of explicitly aligning feature distributions between the two domains. Differently, motivated by the fundamental assumption for domain adaptability, we re-cast the domain adaptation problem as discriminative clustering of target data, given strong privileged information provided by the closely related, labeled source data. Technically, we use clustering objectives based on a robust variant of entropy minimization that adaptively filters target data, a soft Fisher-like criterion, and additionally the cluster ordering via centroid classification. To distill discriminative source information for target clustering, we propose to jointly train the network using parallel, supervised learning objectives over labeled source data. We term our method of distilled discriminative clustering for domain adaptation as DisClusterDA. We also give geometric intuition that illustrates how constituent objectives of DisClusterDA help learn class-wisely pure, compact feature distributions. We conduct careful ablation studies and extensive experiments on five popular benchmark datasets, including a multi-source domain adaptation one. Based on commonly used backbone networks, DisClusterDA outperforms existing methods on these benchmarks. It is also interesting to observe that in our DisClusterDA framework, adding an additional loss term that explicitly learns to align class-level feature distributions across domains does harm to the adaptation performance, though more careful studies in different algorithmic frameworks are to be conducted. %The code is available at \url{https://github.com/huitangtang/DisClusterDA}. 
		\end{abstract}
	\end{frontmatter}
	%\linenumbers
	
	\section{Introduction}
	\label{SecIntro}
	
    Deep learning of neuron networks has made remarkable progress in a wide range of machine learning tasks, with image classification~\cite{ilsvrc} as a prominent example. However, this progress depends heavily on a large amount of labeled data, which are difficult to collect or annotate in many tasks of domains of interest. To address it, we often utilize data in a label-rich source domain to facilitate classification of data in a label-scarce target domain. Nonetheless, there exists a distribution discrepancy between data in the two domains, e.g. synthetic-to-real domain shift~\cite{visda2017}, such that the classifier trained on source samples cannot be reliably applied to target ones. To solve it, a general strategy is domain adaptation~\cite{survey_tl}.

	Given labeled data on a source domain and unlabeled data on a target domain, unsupervised domain adaptation concerns with classification of target data that share a common label space with source data. Assuming a hypothesis space, the classification risk of data on the target domain is theoretically bounded by a combination of three terms~\cite{da_theory1,da_theory2}: the source risk, a measure of discrepancy between distributions of the two domains, and an ideal joint risk that measures adaptability of the given task. %The third term 3) is generally assumed small; otherwise, domain adaptation seems very difficult if not infeasible. 
	%In other words, the discrepancy between the two domains is small enough so that a hypothesis of classifier exists that performs well on both of the two domains. 
	Observing that large-scale deep learning is powerful to obtain features more transferrable across domains and tasks~\cite{imagenet,domain_confusion,how_transferable}, recent domain adaptation methods are designed to align the two domains by learning domain-invariant deep features, such that the second term in the above bound is minimized; representative works include those based on domain-adversarial training~\cite{dann,simnet,ican}. More recently, domain alignment is pushed finer onto the class level to address the ambiguity in class differentiation~\cite{mdd,cdan,tpn,mstn,gsda}, which would appear if the alignment of distributions on whole domains is taken into account only.
	
	%Differently, the basic assumption for domain adaptability~\cite{da_theory1,da_theory2,on_learn_invariant_rep,bsp,tat} (i.e. the third term 3) in the above bound) suggests an alternative direction to pursue, which is to learn classification of target data directly, with no explicit domain-level or class-level feature alignments. Correspondingly, the problem of domain adaptation can be cast as discriminative clustering of target data, given strong privileged information provided by the closely related, labeled source data. In this work, we are motivated to study this alternative paradigm in the context of end-to-end feature and classification learning in a deep network, aiming to push its limit by properly distilling discriminative source information for clustering of target data. 
	
	We note that the third term in \cite{da_theory1,da_theory2} is defined as a sum of the source and target errors predicted by the ideal joint hypothesis, which should be small, i.e., the essential assumption for domain adaptability. We should attach importance to this term since it is not fixed in deep learning with unfixed features, as emphasized in \cite{atda}. Moreover, some recent works aim to reduce it, e.g., \cite{tpn,mstn,pfan,cat}, by explicitly aligning source true centroids and target pseudo ones of the same classes. 
	Intuitively, the definition of this term suggests an alternative direction to pursue, which is to learn classification of target data directly, with no explicit domain-level or class-level feature alignments. Correspondingly, the problem of domain adaptation can be cast as discriminative clustering of target data, given strong privileged information provided by the closely related, labeled source data. In this work, we are motivated to study this alternative paradigm in the context of end-to-end feature and classification learning in a deep network, aiming to push its limit by properly distilling discriminative source information for clustering of target data. \emph{Once the accuracy of pseudo label prediction on target examples is improved, this term is reduced}. 
	On the other hand, the recent arguments in~\cite{bsp,tat,dst_elm} tell that explicit feature alignment (i.e. explicitly modeling and minimizing the domain discrepancy) could hurt the intrinsic discrimination of target data; a recent theoretical work \cite{on_learn_invariant_rep} also tells that minimizing the distribution discrepancy between the source and target features will only increase the target error if the marginal label distributions are significantly different across domains (cf. Theorem 4.3). Our studied paradigm is consistent with these new findings.
	
	Technically, assuming availability of initial cluster assignments, we use clustering objectives based on a robust variant of entropy minimization~\cite{ssl_em,rim} that adaptively filters target samples and favors low-density cluster separation, a soft Fisher-like criterion~\cite{pc_book} that learns deep features by minimizing intra-cluster distances while maximizing inter-cluster ones, and additionally the centroid classification that maintains consistent cluster ordering across domains; benign clustering initialization is enabled by the nature of shared label space between the source and target domains. To distill discriminative source information for classification of target data, we propose to jointly train the network using parallel, supervised learning objectives over the labeled source data. We term our proposed method of distilled discriminative clustering for domain adaptation as \emph{DisClusterDA}. Given shared feature and classification learning across domains, DisClusterDA can be viewed as learning to align the two domains \emph{implicitly}, in contrast to most existing methods~\cite{mdd,tpn,symnet,caada,dmrl} that strive to align feature distributions across domains \emph{explicitly}. %The key novelty of DisClusterDA lies in its \emph{implicit manner} to align the source and target domains, i.e., joint clustering and classification training%, which is consistent with the recent arguments in~\cite{on_learn_invariant_rep,bsp,tat,dst_elm}, which tell that explicit feature alignment (i.e. explicitly modeling and minimizing the domain discrepancy) could hurt the intrinsic discrimination of target data. 
	We also present geometric intuition that illustrates how constituent objectives of DisClusterDA help learn class-wisely pure, compact feature distributions, which are amenable to target classification. 
	
	We conduct careful ablation studies and extensive experiments on five popular benchmark datasets, including a multi-source domain adaptation one. 
	Experiments show the effectiveness of our method; particularly, based on commonly used backbone networks, DisClusterDA outperforms all existing methods on these benchmarks. It is also interesting to observe that in our DisClusterDA framework, adding an additional loss term that explicitly learns to align class-level feature distributions across domains does harm to the adaptation performance. The observation empirically corroborates our motivation in this work, although more careful studies in different algorithmic frameworks are certainly necessary to be conducted. Our main contributions are as follows.
	\begin{itemize}
	\item Motivated by the essential assumption for domain adaptability, we propose to reformulate the domain adaptation problem as discriminative clustering of target data, given strong privileged information from the semantically related, labeled source data. By properly distilling discriminative source information for clustering of target data, we aim to learn classification of target data directly, with no explicit feature alignment. 
	
	\item Technically, we employ clustering objectives based on a robust variant of entropy minimization for reliable cluster separation, a soft Fisher-like criterion for inter-cluster isolation and intra-cluster purity and compactness, and the centroid classification for consistent cluster ordering across domains. To distill discriminative source information for target clustering, we use parallel, supervised learning objectives on the labeled source data. We term our method of distilled discriminative clustering for domain adaptation as \emph{DisClusterDA}. 
	
	\item We also give geometric intuition that illustrates how constituent objectives of DisClusterDA help learn class-wisely pure, compact feature distributions, which are amenable to target classification.
	
	\item Experiments on five widely used benchmark datasets show that our proposed DisClusterDA achieves the new state of the art. %Interestingly, in our DisClusterDA framework, adding the loss term that explicitly learns the class-level feature alignment is bad for target classification, corroborating our motivation in this work, although more careful studies in diverse algorithmic frameworks are certainly necessary to be conducted. 
	\end{itemize}
	
	The rest of this paper is organized as follows. Section \ref{SecRelatedWorks} briefly reviews related works. In Section \ref{SecMethod}, we introduce the proposed method in detail. In Section \ref{SecGeoIntuition}, we analyze our key designs from a geometric perspective. In Section \ref{SecExp}, we show and discuss experimental results. Section \ref{SecConclusion} includes the conclusion and future work.

	\section{Related Works}
	\label{SecRelatedWorks}
	
	%Recent deep unsupervised domain adaptation methods can be grouped into two classes, i.e., homogeneous and heterogeneous settings~\cite{GLG_HeUDA}. In this section, we focus on the more popular homogeneous setting. 
	In this section, we briefly review existing methods from the following three research directions and discuss their relations with our proposed method. One may refer to~\cite{survey_deepVDA} for a comprehensive review of deep visual domain adaptation approaches. 

    \subsection{Explicit Domain Adaptation}
    
    A popular strategy for learning domain-invariant deep features is to explicitly model and minimize distribution discrepancy between the source and target domains. Typically, maximum mean discrepancy (MMD)~\cite{MMD_D,dan,jan,BeyondSW}, graph-matching metric~\cite{graph_alignment,graph_base_alignment}, correlation alignment loss~\cite{caada,deep_coral}, and adversarial training loss~\cite{dann,simnet,ican,mdd,cdan,mstn,gsda,symnet,caada,dmrl,rca,cycada,mada,sbada_gan,mcd,swd,dirt_t,vicatda,ctsn} are used to measure the domain discrepancy. 
    Previous methods~\cite{dann,simnet,ican,dan,BeyondSW} align feature distributions of the source and target domains as a whole, e.g., learning the feature extractor via a reverse signal from the domain classifier that distinguishes between the two domains. 
    Recent methods %\cite{mdd,cdan,tpn,mstn,symnet,caada,rca,mada,mcd,swd,dirt_t,vicatda,gaacn,ctsn,gsda,gvb,dada,dmrl,aada} 
    push the feature alignment from the whole domain level towards finer class level by utilizing discriminative information from the target domain. 
    For example, the methods~\cite{tpn,mstn} assign target data pseudo labels and then enforce overlap between the labeled source and pseudo-labeled target centroids of the same classes. 
    In~\cite{caada}, the distance between feature covariance matrices of the source and target data is minimized to achieve alignment between the same classes from the source and target domains. 
    Discriminative information for both domains is taken into account in~\cite{mdd} to explicitly model a class-aware domain discrepancy, aiming to further achieve the class-level domain alignment. 
    In~\cite{cdan,mada}, the domain classifier~\cite{dann} is redesigned via embedding the multiplicative interaction between instance features and category predictions. 
    Two individual task classifiers are utilized in~\cite{mcd,swd} to detect non-discriminative target features, which are learned to be discriminative by the feature extractor. 
    The methods~\cite{symnet,rca,vicatda} rely on a joint domain-category classifier to learn deep features that are invariant at corresponding classes of the two domains. Tang and Jia~\cite{vicatda} also introduce vicinal domain adaptation, where the vicinal domains are produced by cross-domain mixup. 
    In~\cite{dmrl}, the domain-adversarial training is regularized by both category and domain mixups at the pixel level on individual domains. 
    %Yang \emph{et al.}~\cite{aada} conduct an asymmetric adversarial competition between the feature extractor and an autoencoder-based domain discriminator, where the latter learns to embed the source domain but no the target one and the former learns to deceive the latter by embedding the target domain. 
    %
    %Chen and Hu~\cite{gaacn} propose to embed an attention module in GAN, which allows the discriminator to discriminate transferable regions among images of the two domains. 
    Zuo \emph{et al.}~\cite{ctsn} first identify tough target samples and then handle them using a GAN with two classifiers, which utilizes easy samples and the prediction discrepancy between the two classifiers. 
    Hu \emph{et al.}~\cite{gsda} enforce consistent calibration on both local and global distributions by constraining the gradients of local and global alignments to be synchronous.
    %Cui \emph{et al.}~\cite{gvb} present a gradually vanishing bridge mechanism to model the residual domain-specific characteristics in domain-invariant representations. 
    %
    More recently, Chen \emph{et al.}~\cite{bsp} tell that explicit feature alignment could distort the discriminative structures of target data and thus produce degraded results of target classification; to this end, they penalize the largest singular values of the instance feature matrix, which represent the transferability achieved by explicitly aligning feature distributions between the source and target domains. 
    Differently, with no explicit feature alignment, our proposed DisClusterDA investigates deep discriminative clustering of target data with the help of properly distilled source discriminative information, which can avoid the damage to the discriminative data structures and thus achieve superior performance.
    
    \subsection{Implicit Domain Adaptation}
    
    There have been some attempts at adapting different domains in an implicit manner, yet this direction of research is still largely unexplored. 
    To guarantee adaptability, Liu \emph{et al.}~\cite{tat} abandon feature learning and generate adversarial examples to bridge the domain gap. Saito \emph{et al.}~\cite{atda} train two individual classifiers on both labeled source and pseudo-labeled target data to reduce the domain discrepancy in terms of the disagreement of the two classifiers. 
    Cui \emph{et al.}~\cite{bnm} maximize the batch nuclear-norm on classification response matrix to achieve both discriminability and diversity of feature representations. 
    Some existing non-deep adaptation methods~\cite{dst_elm,dga_da,ldada} are also along this line. For example, Chen \emph{et al.}~\cite{dst_elm} apply an extreme learning machine based space learning algorithm for domain space transfer. Lu \emph{et al.}~\cite{ldada} propose a linear-discriminant-analysis-like framework to learn class-specific linear projections by only using the class mean; they regard the source and target domains as one domain, where the original Fisher's criterion is exactly applied. In contrast, we parallelly apply a soft Fisher-like criterion to the individual domain of source and target to enable source distilled discriminative target clustering.
    
    \subsection{Deep Discriminative Clustering}
    
    Classical methods for discriminative clustering~\cite{rim,adapt_dim_reduct,discriminat_kmeans} follow the Fisher-style criteria, and alternate between subspace learning and data clustering. Recent deep methods~\cite{dc_ul_vf,depict,dc_on_the_link,dec} simultaneously learn feature representation and cluster assignment via end-to-end network training, which are based on the representative clustering principles, e.g., mutual information~\cite{dc_on_the_link} or Kullback-Leibler (KL) divergence~\cite{depict,dec}. 
    Differently, we have a closely related source domain that enables us to exploit the full Fisher's criterion, which is largely absent in existing discriminative clustering algorithms. Labeled instances in the source domain serve as a sort of privileged information~\cite{unify_distill_privilege,learn_use_privilege}. The discriminative source information is transferred to clustering of target data via joint network training, which is essentially a form of distillation~\cite{distill_knowledge}. 
    However, our DisClusterDA is completely different from those knowledge distillation based domain adaptation methods; it does not do the model distillation \cite{kd_for_ss_da,da_task_distill} but distills the source domain knowledge by minimizing the sum of classification, Fisher, and ordering losses, rather than selecting source samples \cite{msda_distill}. 
    Moreover, these methods have not utilized the clustering algorithm. 
    We note that some domain adaptation methods~\cite{cat,rca,dirt_t,vicatda,larger_norm,learn_to_clust} have used the typical clustering techniques. For example, Shu \emph{et al.}~\cite{dirt_t} constrain the domain-adversarial training~\cite{dann} with entropy minimization~\cite{min_ent}. 
    Deng~\emph{et al.}~\cite{cat} borrow from Smooth Neighbors on Teacher Graphs (SNTG)~\cite{sntg}, which considers the spatial proximity between any two data points. 
    %Kang \emph{et al.}~\cite{can} apply spherical k-means clustering~\cite{spherical_k_means} to generate more reliable pseudo labels for target samples. 
    Hsu~\emph{et al.}~\cite{learn_to_clust} learn to cluster by transferring pairwise semantic similarity. 
    However, they still force explicit domain alignment. 
    In contrast, we propose a novel \emph{implicit manner} to align the source and target domains, i.e. joint clustering and classification training; our idea of performing clustering elegantly integrates adaptive filtering entropy minimization, soft Fisher-like criterion, and cluster ordering via centroid classification, which can derive more sensible clustering solutions.

    \begin{figure}[!t]
    	\centering
    	\includegraphics[width=1.0\textwidth]{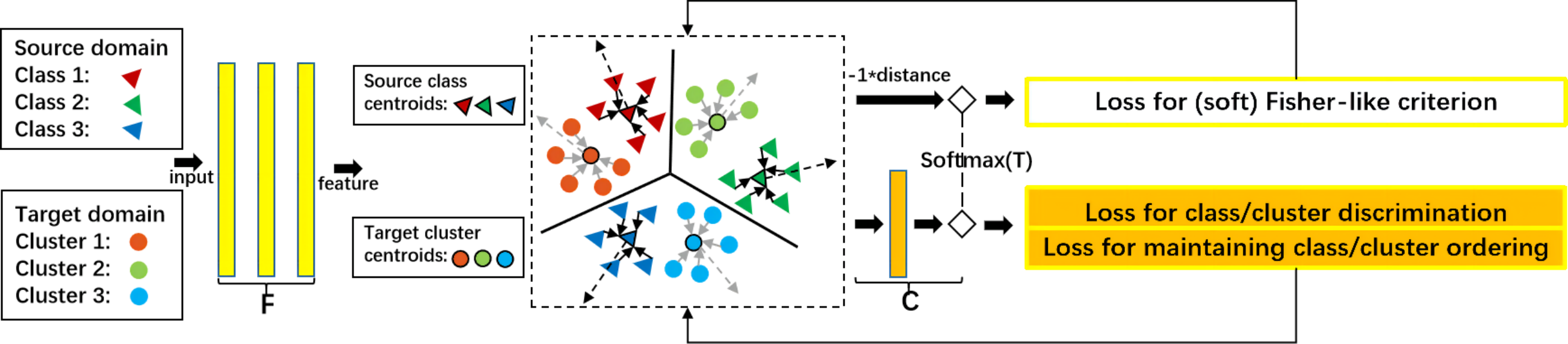}	 
    	\caption{Overall network and objective of our proposed DisClusterDA. The proposed losses are minimized over a feature extractor $F$ and a task classifier $C$, effects of which are shown in the middle dashed rectangle. %Best viewed in color.
    	}
    	\label{fig:our_method}%\vspace{-0.1cm}
    \end{figure}

	\section{Distilled Discriminative Clustering}
	\label{SecMethod}
	
	In unsupervised domain adaptation, we are given labeled source instances $\{(\mathbf{x}_i^s, y_i^s)\}_{i=1}^{n_s}$ and unlabeled target instances $\{\mathbf{x}_j^t\}_{j=1}^{n_t}$, which are respectively sampled from the source domain $({\cal{D}}_s^X, {\cal{Y}})$ and target one ${\cal{D}}_t^X$. Let $|{\cal{Y}}| = K$, and we have $y \in \{1, \dots, K\}$ for any instance $\mathbf{x}$. As illustrated in Fig.~\ref{fig:our_method}, we study domain adaptation in the context of end-to-end learning a deep network that stacks a task classifier $C: {\cal{F}} \rightarrow \mathbb{R}^K$, followed by softmax operation, on top of a feature extractor $F: {\cal{D}}^{X} \rightarrow {\cal{F}}$, where ${\cal{F}}$ denotes the deep feature space. We write $\mathbf{f} = F(\mathbf{x}) \in {\cal{F}}$ and $\mathbf{z} = C(\mathbf{f}) \in \mathbb{R}^K$, and have the probability simplex of the network output as $\mathbf{p} = \mathbf{\sigma}(\mathbf{z}) \in [0, 1]^K$, via the softmax function $\mathbf{\sigma}(\cdot)$. We also write $p_k$ to denote the $k^{th}$ element probability of $\mathbf{p}$. As discussed in Section \ref{SecIntro}, we cast the problem of interest as learning from $\{\mathbf{x}_j^t\}_{j=1}^{n_t}$ a parametric function $C \circ F$ that maps data on ${\cal{D}}_t^X$ into $K$ clusters/classes, given strong privileged information in $\{(\mathbf{x}_i^s, y_i^s)\}_{i=1}^{n_s}$ yet to be discovered. Compared with supervised learning, training network $C \circ F$ via unsupervised clustering produces probability prediction $\mathbf{p}$ of lower confidence, which is modeled in this work with incorporation of a temperature $T$ into the softmax function, i.e.,
	\begin{equation}\label{EqnSoftMaxTemperature}
	p_k = \sigma_{k,T}(\mathbf{z}) = \frac{ \exp(z_k/T) }{ \sum_{k' = 1}^{K} \exp(z_{k'}/T) }. 
	\end{equation}
	Since $\{(\mathbf{x}_i^s, y_i^s)\}_{i=1}^{n_s}$ would be used to train the same network jointly, this is essentially a form of distillation~\cite{distill_knowledge} that transfers discriminative information in the labeled source data to target clustering.
	
	\subsection{Unsupervised Domain Adaptation as Deep Discriminative Clustering}
	
	We present in this section our objective choices that train the network in favor of discriminative clusters. For unlabeled target data $\{ \mathbf{x}_j^t \}_{j=1}^{n_t}$, we maintain dynamic cluster assignments $\{ \hat{y}_j^t \}_{j=1}^{n_t}$ during network training. Cluster assignment $ \hat{y}^t \in \{1, \dots, K\} $ of any $\mathbf{x}^t$ is obtained by $\hat{y}^t  = {\arg\max}_k p_k(\mathbf{x}^t)$~\cite{pseudo_label,ufl_discriminat_encod}. % {\color{red} --- better alternative exists in the context of domain adaptation, as explained in Section \ref{XXX}.} 
	Corresponding to these cluster assignments are the $K$ cluster centroids $\{ \mathbf{m}_k^t \in {\cal{F}} \}_{k=1}^K$ in the deep feature space, which are updated per iteration of network training. Specifically, the centroid for any $k^{th}$ cluster is updated according to the rule of moving average~\cite{mstn}
	\begin{equation}\label{EqnTargetCentroidMA}
	\mathbf{m}_k^t \leftarrow \alpha \mathbf{m}_k^t + (1 - \alpha) \frac{1}{|\tilde{\cal{J}}_k^t|} \sum_{ j \in \tilde{\cal{J}}_k^t } \mathbf{f}_j^t, 
	\end{equation}
	where $\tilde{\cal{J}}_k^t = \{j | \hat{y}_{j}^t = k \}$ denotes the set of target instances in a mini-batch that are assigned to the $k^{th}$ cluster, and $\alpha \in [0, 1]$ is the moving average coefficient.
	
	\paragraph{Adaptive Filtering Entropy Minimization} Our first objective follows entropy regularization~\cite{ssl_em,rim}; it is argued that minimizing conditional entropy of class probabilities captures cluster assumption~\cite{ssc_lds}, i.e., decision boundaries of models should locate in regions of lower density, thus achieving cluster discrimination. On the other hand, established studies in cluster analysis have shown that clusters can be estimated with a low probability of error only on condition of small conditional entropy~\cite{fano_ineq}. To improve over~\cite{ssl_em}, we propose an adaptive filtering variant of entropy loss as
	\begin{eqnarray}\label{EqnLossTargetWeightedEM}
	\begin{aligned}
	{\cal{L}}_{entropy}^t  (F, C) = \frac{1}{n_t}\sum_{j=1}^{n_t} \mathrm{e}^{ - {\cal{H}}\left( \mathbf{\sigma}_T\left( C\circ F(\mathbf{x}_j^t) \right) \right) } {\cal{H}}\left( \mathbf{\sigma}_T\left( C\circ F(\mathbf{x}_j^t) \right) \right), 
	\end{aligned}
	\end{eqnarray}
	where for a probability vector $\mathbf{p}$, its entropy is computed as $ {\cal{H}}(\mathbf{p}) = - \sum_{k=1}^K p_k\log p_k$. During network training, any instance $\mathbf{x}^t$ with relatively even predictions of element probabilities $\{ p_k(\mathbf{x}^t) \}_{k=1}^K$ is less confident about its cluster assignment, the value of its entropy ${\cal{H}}(\mathbf{p}(\mathbf{x}^t))$ is higher, and its importance is thus weighted down in (\ref{EqnLossTargetWeightedEM}) by exponential function of its negative entropy. 
	More specifically, given that the task at hand has $K$ classes, the maximum of entropy is $\log K$ and the minimum of negative entropy is $-\log K$; the lower bound of exponent of negative entropy is thus $e^{-\log K}$, e.g., $0.032$ when $K=31$. It makes sense that the lower bound of exponent of negative entropy is smaller with more categories, since the lower bound of confidence ($1/K$) is also smaller; hence, the negative effects caused by the less confident instances are weighted more down. %Our used exponential weighting has the similar insight and effect as pseudo-labeling \cite{fixmatch} that uses a pre-defined confidence threshold to select reliable target samples. If the confidence of a sample is low (i.e. its entropy is large), its exponent of negative entropy is close to $0$, i.e. not selected. 
	Note that we focus more on high-confidence instances in a way similar to the hard-thresholded pseudo-labeling \cite{fixmatch}, but more smoothly and flexibly; the instances with low confidence still contribute to the model training, except those with a lower degree. 
	We emphasize that there does not exist a fixed/pre-defined threshold in our adaptive filtering entropy minimization, and it means that we avoid the introduction of an additional hyper-parameter. 
	Our used exponential weighting naturally admits such a filtering behavior in a more reasonably adaptive manner. 
	Objective (\ref{EqnLossTargetWeightedEM}) thus achieves improved robustness by relying on samples with more confident predictions in earlier stages of clustering. This also complies with the optimization dynamics of supervised learning~\cite{CloserLookDN} that first learns easier samples that better fit patterns.
	
	Minimizing conditional entropy alone suffers from degenerate solutions of removed clusters and decision boundaries; as a remedy, a term in favor of class balance is commonly used to have mutual information based discriminative clustering~\cite{rim,dc_on_the_link}. Class imbalance is not a critical issue in the task setting of domain adaptation; we instead propose the following soft Fisher-like criterion to enhance discrimination between clusters.
	
	\paragraph{Soft Fisher-like Criterion} To enhance discrimination in the deep feature space ${\cal{F}}$, we augment (\ref{EqnLossTargetWeightedEM}) via a soft Fisher-like criterion that learns deep features such that intra-cluster distances are minimized, and inter-cluster ones are maximized. For any $\mathbf{x}^t$ with its feature $\mathbf{f}^t$, denote its squared distance vector w.r.t. the $K$ centroids $\{ \mathbf{m}_k^t \}_{k=1}^K$ as $\mathbf{d}_w^t = [\dots, - \| \mathbf{f}^t - \mathbf{m}_k^t \|_2^2, \dots]^{\top} \in \mathbb{R}_{\leq 0}^K$; for any $k^{th}$ centroid $\mathbf{m}_k^t$, denote its squared distance vector w.r.t. $\{ \mathbf{m}_k^t \}_{k=1}^K$ (including itself) as $ \mathbf{d}_{b,k}^t = [\dots, - \| \mathbf{m}_k^t - \mathbf{m}_{k-1}^t \|_2^2, 0,  - \| \mathbf{m}_k^t - \mathbf{m}_{k+1}^t \|_2^2, \dots]^{\top} \in \mathbb{R}_{\leq 0}^K$. We technically achieve a soft Fisher-like criterion using again the adaptive filtering entropy formulation
	\begin{align}\label{EqnLossTargetFisher}
	%\begin{aligned}
	{\cal{L}}_{Fisher}^t & (F)  = \frac{1}{n_t}  \sum_{j=1}^{n_t}  \mathrm{e}^{-  {\cal{H}}\left( \mathbf{\sigma}_T  \left( \mathbf{d}_{w,j}^t(F) \right) \right) } {\cal{H}}  \left( \mathbf{\sigma}_T  \left( \mathbf{d}_{w,j}^t(F) \right) \right) \nonumber \\ &  + \frac{1}{K}  \sum_{k=1}^K  \mathrm{e}^{-  {\cal{H}}\left( \mathbf{\sigma}_T  \left( \mathbf{d}_{b,k}^t(F) \right) \right) } {\cal{H}}  \left( \mathbf{\sigma}_T  \left( \mathbf{d}_{b,k}^t(F) \right) \right). 
	%\end{aligned}
	\end{align}
	The first term in (\ref{EqnLossTargetFisher}) is to push feature $\mathbf{f}^t$ of any instance $\mathbf{x}^t$ closer to one of the $K$ centroids, with consideration of its current distance-based confidence level; the second term in (\ref{EqnLossTargetFisher}) is to repulse any centroid $\mathbf{m}_k^t$ away from the other $K-1$ ones, with consideration of the current level of centroids' distinctiveness. Fisher-style criteria are adopted in classical discriminative clustering methods~\cite{adapt_dim_reduct,discriminat_kmeans}, where subspace learning and clustering are conducted alternately; in a deep network instead, feature learning and clustering are conducted simultaneously via stochastic end-to-end training.
	
	\paragraph{Cluster Ordering via Centroid Classification} The objectives (\ref{EqnLossTargetWeightedEM}) and (\ref{EqnLossTargetFisher}) enforce soft assignments of $\{\mathbf{x}_j^t\}_{j=1}^{n_t}$ into $K$ clusters; however, ordering of these clusters is determined up to arbitrary permutations, which causes inconvenience of distillation via jointly training the network on source data. With no loss of generality, we assume the initial ordering of the $K$ clusters is aligned with the $K$ output neurons of the network. We use cross-entropy loss on centroids $\{ \mathbf{m}_k^t \}_{k=1}^K$ to maintain the ordering
	\begin{equation}\label{EqnTargetCentroidClassification}
	{\cal{L}}_{ordering}^t (F, C) = - \frac{1}{K} \sum_{k=1}^K \log \sigma_{k, T}\left( C\left( \mathbf{m}_k^t(F) \right) \right) .
	\end{equation}
	Objective (\ref{EqnTargetCentroidClassification}) has additional and important benefit of improving continuity, purity, and compactness of cluster-wise feature distributions in ${\cal{F}}$. We geometrically explain its importance in Section \ref{SecGeoIntuition}.
	
	\paragraph{Overall Clustering Objective} Combining the objectives (\ref{EqnLossTargetWeightedEM}), (\ref{EqnLossTargetFisher}), and (\ref{EqnTargetCentroidClassification}) gives the overall objective for deep discriminative clustering
	\begin{equation}\label{EqnTargetOverallObj}
	\begin{aligned}
	{\cal{L}}_{clustering}^t (F, C) = {\cal{L}}_{entropy}^t (F, C) + {\cal{L}}_{Fisher}^t (F) + {\cal{L}}_{ordering}^t (F, C). 
	\end{aligned}
	\end{equation}
	Objective (\ref{EqnTargetOverallObj}) is related to those of recent deep discriminative clustering methods~\cite{dc_ul_vf,depict,dc_on_the_link,dec}. These methods typically employ an additional reconstruction loss to prevent overfitting of deep feature learning; we note that in domain adaptation, this issue is avoided by jointly training the same network on labeled source data. Our soft formulations (\ref{EqnLossTargetWeightedEM}) and (\ref{EqnLossTargetFisher}) may alleviate this issue as well.

	\subsection{Distilling Discriminative Source Information via Joint Network Training}
	
	The fundamental assumption for the feasibility of domain adaptation~\cite{da_theory1} suggests that it is possible to learn a classifier $C$ to perform well on both ${\cal{D}}_s^X$ and ${\cal{D}}_t^X$, even when they are not aligned in the feature space ${\cal{F}}$. This motivates us to train the same network $C \circ F$ on the labeled source data $\{(\mathbf{x}_i^s, y_i^s)\}_{i=1}^{n_s}$ as well, such that the learned $C$ can discriminate both the source and target classes/clusters. Technically, we propose the following parallel, supervised learning objectives over $\{(\mathbf{x}_i^s, y_i^s)\}_{i=1}^{n_s}$ to jointly train the network
	\begin{eqnarray}\label{EqnSrcOverallObj}
	\begin{aligned}
	{\cal{L}}_{distilling}^s (F, C) =  {\cal{L}}_{cls}^s (F, C) + {\cal{L}}_{Fisher}^s (F) + {\cal{L}}_{ordering}^s (F, C). 
	\end{aligned}
	\end{eqnarray}
	The constituent objectives of (\ref{EqnSrcOverallObj}) are in parallel with (\ref{EqnLossTargetWeightedEM}), (\ref{EqnLossTargetFisher}), and (\ref{EqnTargetCentroidClassification}), and are respectively defined as
	\begin{eqnarray}\label{EqnLossSrcClassification}
	\begin{aligned}
	{\cal{L}}_{cls}^s(F, C)  =  - \frac{1}{n_s}\sum_{i=1}^{n_s} \log \sigma_{y_i^s, T}\left( C\circ F(\mathbf{x}_i^s) \right) ,
	\end{aligned}
	\end{eqnarray}
	
	\begin{eqnarray}\label{EqnLossSrcFisher}
	\begin{aligned}
	{\cal{L}}_{Fisher}^s(F) = - \frac{1}{n_s}\sum_{i=1}^{n_s}  \log \sigma_{y_i^s, T}\left( \mathbf{d}_{w, i}^s(F) \right) - \frac{1}{K} \sum_{k=1}^K \log \sigma_{k, T} \left( \mathbf{d}_{b, k}^s(F) \right),
	\end{aligned}
	\end{eqnarray}
	
	\begin{equation}\label{EqnSrcCentroidClassification}
	\begin{aligned}
	{\cal{L}}_{ordering}^s (F, C) = - \frac{1}{K} \sum_{k=1}^K \log \sigma_{k, T}\left( C\left( \mathbf{m}_k^s(F) \right) \right), 
	\end{aligned}
	\end{equation}
	where $\{\mathbf{m}_k^s\}_{k=1}^K$ are class centroids of source features $\{\mathbf{f}_i^s\}_{i=1}^{n_s}$, which are also updated according to (\ref{EqnTargetCentroidMA}), using true labels $\{y_i^s\}_{i=1}^{n_s}$; for any $\mathbf{x}^s$ with its feature $\mathbf{f}^s$, the intra-class distance vector is defined as $\mathbf{d}_w^s = [\dots, - \| \mathbf{f}^s - \mathbf{m}_k^s \|_2^2, \dots]^{\top} \in \mathbb{R}_{\leq 0}^K$, and the inter-class distance vector for any $\mathbf{m}_k^s$ is defined as $ \mathbf{d}_{b,k}^s = [\dots, - \| \mathbf{m}_k^s - \mathbf{m}_{k-1}^s \|_2^2, 0,  - \| \mathbf{m}_k^s - \mathbf{m}_{k+1}^s \|_2^2, \dots]^{\top} \in \mathbb{R}_{\leq 0}^K$; we use cross-entropy losses in (\ref{EqnLossSrcClassification}) and (\ref{EqnLossSrcFisher}), instead of entropy based ones as in (\ref{EqnLossTargetWeightedEM}) and (\ref{EqnLossTargetFisher}), as the labels are available for source instances.
	
	For clustering of $\{\mathbf{x}_j^t\}_{j=1}^{n_t}$, labeled source instances $\{(\mathbf{x}_i^s, y_i^s)\}_{i=1}^{n_s}$ serve as a sort of privileged information~\cite{unify_distill_privilege,learn_use_privilege}. However, in contrast to standard forms that usually specify explanations of $\{\mathbf{x}_j^t\}_{j=1}^{n_t}$, the privileged information in $\{(\mathbf{x}_i^s, y_i^s)\}_{i=1}^{n_s}$ is to be distilled by training the network $C\circ F$ via (\ref{EqnSrcOverallObj}), and would be encoded into network parameters. To use the distilled information for domain adaptation, we propose the following objective of \emph{DisClusterDA} that combines the distillation term (\ref{EqnSrcOverallObj}) and discriminative clustering term (\ref{EqnTargetOverallObj}) for the case that the classifier $C$ is implemented as one fully-connected (FC) layer, resulting in 
	\begin{equation}\label{EqnOverallObj}
	\begin{aligned}
	{\cal{L}}_{DisClusterDA} (F, C) = {\cal{L}}_{distilling} (F, C) + \lambda {\cal{L}}_{clustering} (F, C), 
	\end{aligned}
	\end{equation}
	where $\lambda \in [0,1]$ is to suppress the noisy signal of (\ref{EqnTargetOverallObj}) in the early stage of joint training.
	
	Most of recent deep adaptation methods~\cite{tpn,mstn,gsda,dmrl,mcd} strive to align source and target instances of the same classes in ${\cal{F}}$ \emph{explicitly}. Different from them, our proposed DisClusterDA does not enforce explicit alignment; instead, the labeled source instances are mainly used as structural constraints to regularize discriminative clustering of the target ones. To some extent, DisClusterDA shares a similar insight with the recent shallow method~\cite{dga_da}, which takes the geometric structure of the underlying data manifold into account. Given shared learning of $F$ and $C$, our method can thus be viewed as learning to align the two domains \emph{implicitly}, for which we give geometric intuition shortly. We note that an optional term that minimizes distances $\| \mathbf{m}_k^s - \mathbf{m}_k^t \|_2^2$, $k=1, \dots, K$, of corresponding centroids may be included in our method. Empirical results in Section \ref{SecExp} show that such a scheme produces degraded performance, corroborating our motivation for DisClusterDA, though more careful studies in other algorithmic frameworks are to be investigated.

	\subsection{Network Training and Test}
	
	We summarize the training process of DisClusterDA in Algorithm 1 in the appendix. We train the network with Stochastic Gradient Descent (SGD). In the inference phase, we use the learned network $C \circ F$ to classify any target test sample by $\hat{y}^t  = {\arg\max}_k p_k(\mathbf{x}^t)$. The classification accuracy is calculated as $accuracy = |\mathbf{x}^t: \mathbf{x}^t \in {\mathcal{T}}_{test} \land \hat{y}^t = y^t| / |\mathbf{x}^t: \mathbf{x}^t \in {\mathcal{T}}_{test}|$.

\section{Geometric Intuition}
\label{SecGeoIntuition}

	In this section, we present geometric intuition that illustrates effects achieved by constituent objectives of DisClusterDA for classification of target instances. Our analysis stands on the deep feature space ${\cal{F}}$, which is lifted up via $F$ from the input space ${\cal{D}}^X$. Without loss of generality, we assume ${\cal{F}} \in \mathbb{R}^m$.
	
\begin{figure}[!t]
		\centering
		\subfigcapskip=-8pt
		\subfigure[One FC layer]{\includegraphics[height=1.5in]{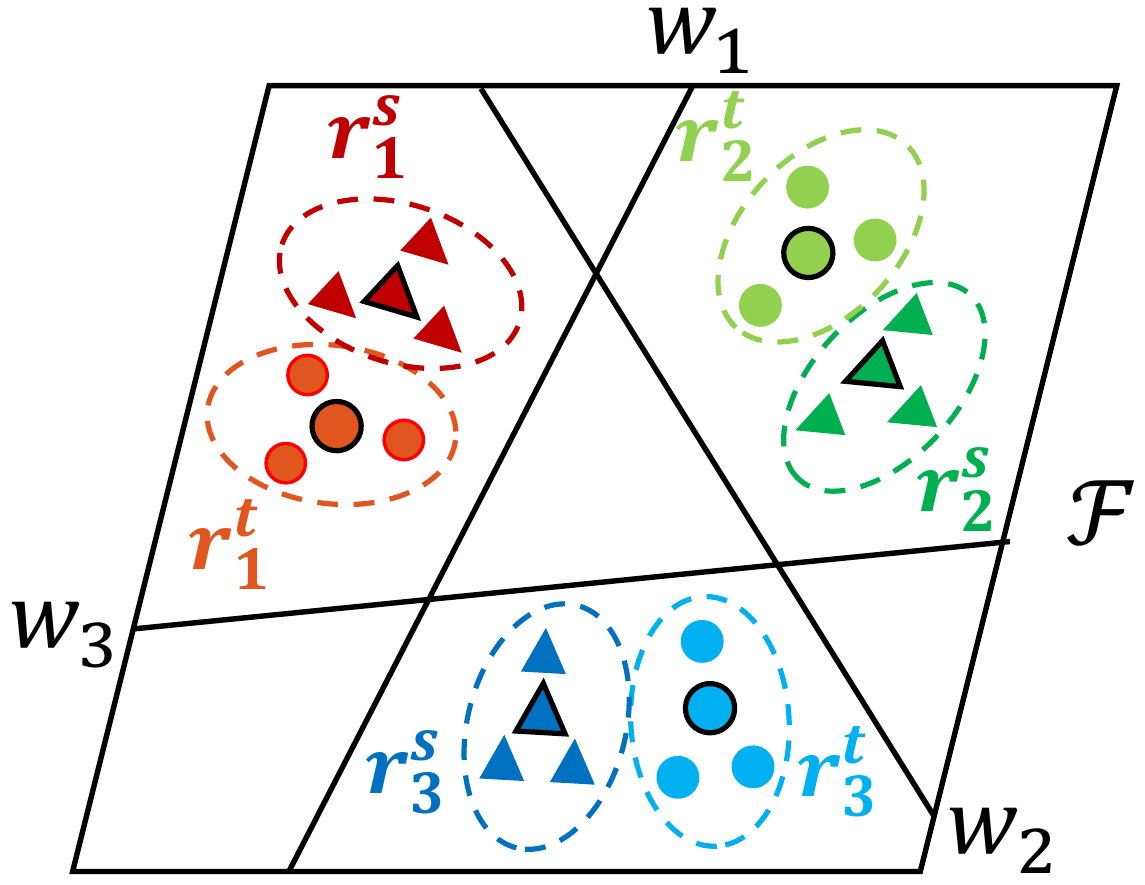}
			\label{fig:geo_exp1}}
		\hspace{0.2in}
		\subfigure[Multiple FC layers]{\includegraphics[height=1.5in]{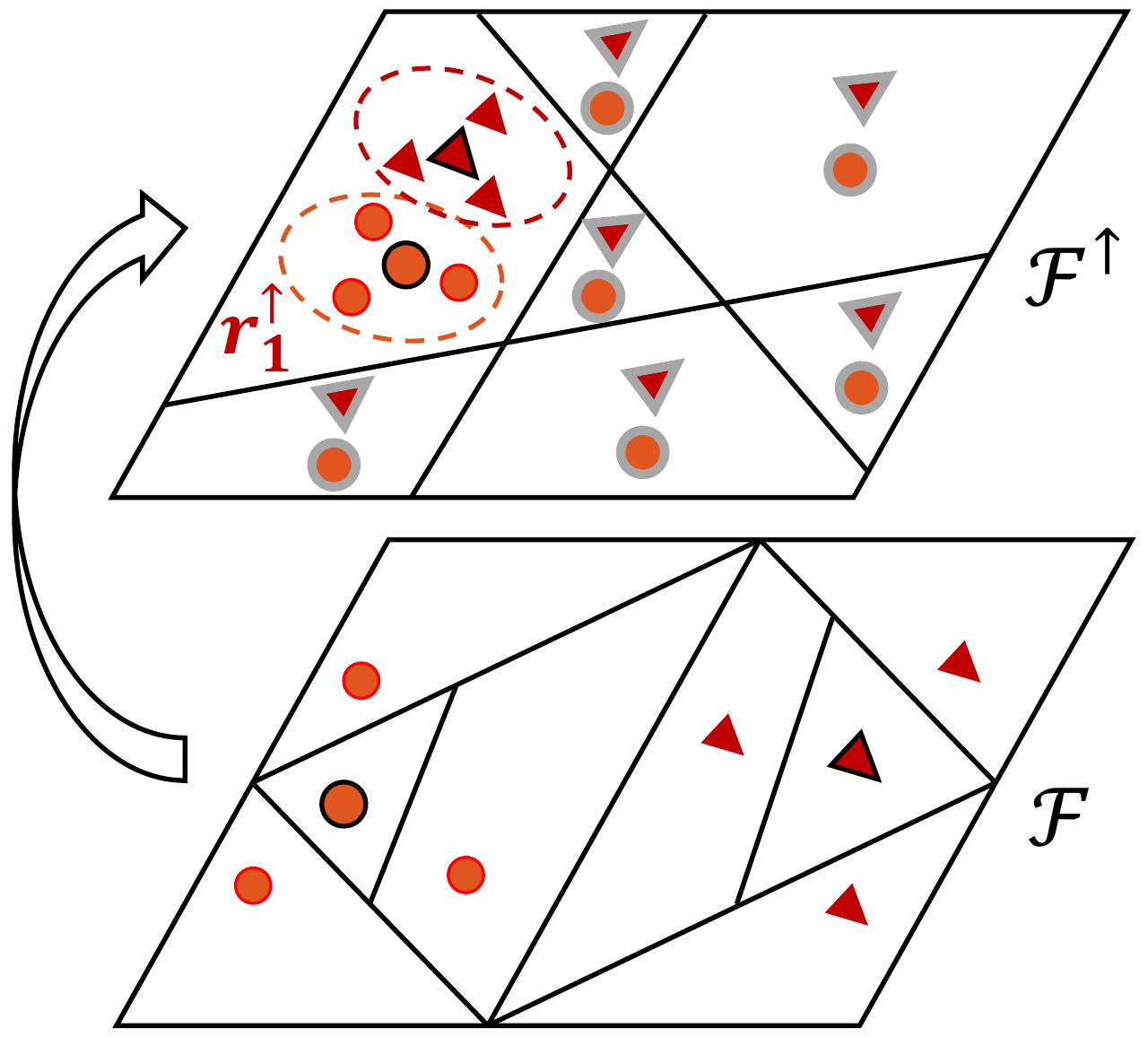}
			\label{fig:geo_exp2}}
%		\begin{minipage}[c]{0.49\textwidth}
%			%\centering
%			\centerline{\includegraphics[height=1.5in]{images/illu_geo_1fc.pdf}} \vspace{-0.3cm}
%			\centerline{(a) One FC layer}
%		\end{minipage}
%		\hspace{0.03in}
%		\begin{minipage}[c]{0.49\textwidth}
%			%\centering
%			\centerline{\includegraphics[height=1.5in]{images/illu_geo_2fcs.pdf}} \vspace{-0.3cm}
%			\centerline{(b) Multiple FC layers}
%		\end{minipage}
		\vspace{-0.3cm}
		\caption{Illustration of effects of our ordering losses (\ref{EqnTargetCentroidClassification}) and (\ref{EqnSrcCentroidClassification}) when the classifier $C$ is implemented as one or multiple FC layers, where triangles and circles with a gray outline are wrongly positioned source and target class centroids, when they are lifted up from $\cal{F}$ to ${\cal{F}}_{\uparrow}$. %Best viewed in color.
		}
		\label{fig:geo_exp}
	\end{figure}

	We first consider the case that the classifier $C$ is implemented as one FC layer. Denote its parameterization as $C(\mathbf{f}) = \mathbf{W}^{\top}\mathbf{f} \in \mathbb{R}^K$, where $\mathbf{W} \in \mathbb{R}^{m\times K}$. We also write the column vectors of $\mathbf{W}$ as $\{ \mathbf{w}_k \}_{k=1}^K$, which in fact specify a hyperplane arrangement~\cite{arrange_hp}, denoted as ${\cal{A}}$ --- a finite hyperplane arrangement is a finite set of affine hyperplanes in some vector space (i.e., $\mathbb{R}^m$). Let $\mathbf{w} \in \mathcal{A}$ be an element of the arrangement. We define a region $r$ as a connected component of the complement $\mathbb{R}^m - \bigcup\limits_{\mathbf{w} \in \mathcal{A}}\mathbf{w}$, and denote all the regions collectively as ${\cal{R}}$. Let $\mathbb{I} = \{1, 0\}$, and define a map $\mathbf{\tau}: \mathcal{F} \rightarrow \mathbb{I}^K$ by
	\begin{displaymath}
	\tau_k(\mathbf{f}) =
	\begin{cases}
	1 & \text{if $ \mathbf{w}_k^{\top} \mathbf{f}  > 0$}, \\
	0 & \text{if $ \mathbf{w}_k^{\top} \mathbf{f} \leq 0$}.
	\end{cases}
	\end{displaymath}
	Any region $r \in {\cal{R}}$ can thus be indexed by a unique element in $\mathbb{I}^K$, denoted as $\mathbf{\tau}(r)$. For any target instance $\mathbf{x}^t$ with its feature $\mathbf{f}^t$, the adaptive filtering entropy term (\ref{EqnLossTargetWeightedEM}) promotes uneven predictions among $\{ \sigma_k(\mathbf{W}^{\top}\mathbf{f}^t)\}_{k=1}^K$; ideally one of them would approach the value of $1$, and the others would approach $0$. This translates as the fact that the objective (\ref{EqnLossTargetWeightedEM}) would drive clustering of $\{ \mathbf{x}_j^t \}_{j=1}^{n_t}$ into $K$ distinct regions $\{r_k^t \in {\cal{R}}\}_{k=1}^K$ in ${\cal{F}}$, whose indexes are given by $\{ \mathbf{\tau}(r_k^t) \in \mathbb{I}^K \}_{k=1}^K$. The classification loss (\ref{EqnLossSrcClassification}) for labeled source instances $\{(\mathbf{x}_i^s, y_i^s)\}_{i=1}^{n_s}$ has the same effect of learning the corresponding $\{ \mathbf{f}_i^s \}_{i=1}^{n_s}$ into $K$ distinct regions $\{r_k^s \in {\cal{R}}\}_{k=1}^K$ indexed by $\{ \mathbf{\tau}(r_k^s) \in \mathbb{I}^K \}_{k=1}^K$. Enforcing ordering of the $K$ clusters to be aligned with output neurons of the classifier $C$ geometrically means that $\mathbf{\tau}(r_k^s) = \mathbf{\tau}(r_k^t)$, for $k = 1, \dots, K$, as shown in Fig.~\ref{fig:geo_exp} (a).
	
	Now consider the case that $C$ is implemented as a subnetwork consisting of multiple FC layers, for which the overall objective is presented in the appendix. Denote input space of the last FC layer of $C$ as ${\cal{F}}^{\uparrow}$, and its induced region space as ${\cal{R}}^{\uparrow}$. While the objectives (\ref{EqnLossTargetWeightedEM}) and (\ref{EqnLossSrcClassification}) would cluster or classify the target and source instances into $K$ distinct regions $\{ r^{\uparrow}_{k} \in {\cal{R}}^{\uparrow} \}_{k=1}^K$, indexed by $\{ \mathbf{\tau}(r^{\uparrow}_k) \in \mathbb{I}^K \}_{k=1}^K$, in the space ${\cal{F}}^{\uparrow}$, instances belonging to the same region $r^{\uparrow}_{k}$ are not necessarily positioned in a same region $r \in {\cal{R}}$. In fact, due to the \emph{space folding} effect of deep networks~\cite{linear_region}, they would be fragmented in different regions in ${\cal{R}}$. Consequently, we may have ${\mathbf{m}}_k^{t\uparrow} \not\in r^{\uparrow}_k$ and ${\mathbf{m}}_k^{s\uparrow} \not\in r^{\uparrow}_k$ for target and source centroids $\mathbf{m}_k^t$ and $\mathbf{m}_k^s$ of the $k^{th}$ cluster/class, when they are lifted up to the space ${\cal{F}}^{\uparrow}$, as illustrated in Fig.~\ref{fig:geo_exp} (b). Our use of the objectives (\ref{EqnTargetCentroidClassification}) and (\ref{EqnSrcCentroidClassification}) for centroid classification would have the effect of encouraging continuity and purity of class-wise feature distributions in ${\cal{F}}$, by learning to enforce ${\mathbf{m}}_k^{t\uparrow} \in r^{\uparrow}_k$ and ${\mathbf{m}}_k^{s\uparrow} \in r^{\uparrow}_k$.
	
	The Fisher-like objectives (\ref{EqnLossTargetFisher}) and (\ref{EqnLossSrcFisher}) further enhance the intra-class purity and compactness and inter-class distinctiveness of feature distributions in ${\cal{F}}$. To illustrate geometrically, denote the angle between an instance $\mathbf{f}^t$ and its centroid $\mathbf{m}_k^t$ of the $k^{th}$ cluster in ${\cal{F}}$ as $\theta_w^t (\mathbf{f}^t, \mathbf{m}_k^t) = \cos^{-1} ( \mathbf{m}_k^{t\top} \mathbf{f}^t / \| \mathbf{m}_k^t \|_2 \| \mathbf{f}^t \|_2 ) $, and define the angle between any two centroids of different clusters as $\theta_b^t (\mathbf{m}_k^t, \mathbf{m}_{k'}^t) = \cos^{-1} ( \mathbf{m}_k^{t\top} \mathbf{m}_{k'}^t / \| \mathbf{m}_k^t \|_2 \| \mathbf{m}_{k'}^t \|_2 ) $. The objective (\ref{EqnLossTargetFisher}) aims to minimize any $\theta_w^t$, while maximizing any $\theta_b^t$; objective (\ref{EqnLossSrcFisher}) has the same effect for $\theta_w^s$ and $\theta_b^s$.
	
	In our method, the classifier $C$ is shared for both domains, which is optimized by target clustering and source classification simultaneously, since the source domain has ground-truth labels and the target one does not. Note that the $K$-way classifier $C$ defines hyperplanes that partition the feature space ${\cal{F}}$ into regions whose number is bounded by $2^K$, and $K$ ones among them are uniquely responsible for the $K$ classes \cite{linear_region}. Given that the two domains have the same label space, joint clustering and classification training, especially the proposed cluster ordering via centroid classification, would \emph{ideally} push instances of the two domains from a specific class into the same region in ${\cal{F}}$, thus \emph{implicitly} achieving adaptation between the two domains.%, as discussed in Section 4 in the previous manuscript.

\section{Experiments}
\label{SecExp}

\subsection{Datasets}

\textbf{Office-31}~\cite{office31} is a popular benchmark dataset for visual domain adaptation, which contains $4,110$ images of $31$ classes from three different domains: Amazon (\textbf{A}) which includes images downloaded from \url{amazon.com}, Webcam (\textbf{W}) and DSLR (\textbf{D}) which include images respectively taken by web camera and digital SLR camera under different settings. We follow a common protocol and evaluate on $6$ adaptation tasks.

\textbf{Office-Home}~\cite{officehome} is a difficult benchmark dataset, which collects about $15,500$ images of $65$ object classes from office and home scenes, forming four extremely distinct domains: Artistic (\textbf{Ar}), Clip Art (\textbf{Cl}), Product (\textbf{Pr}), and Real-World (\textbf{Rw}). We evaluate on $12$ adaptation tasks formed by combining any two domains.

\textbf{Digits} includes three domains: SVHN (\textbf{S})~\cite{svhn} which contains $99,289$ RGB images where more than one digit may exist, MNIST (\textbf{M})~\cite{lenet} which comprises $70,000$ grayscale images with clean background, and USPS (\textbf{U})~\cite{usps} which is composed of $9,298$ grayscale images with unconstrained writing styles, making a good complement to the above two datasets for diverse domain adaptation scenarios. We follow tradition and evaluate on $4$ adaptation tasks: \textbf{M}$\rightarrow$\textbf{S}, \textbf{S}$\rightarrow$\textbf{M}, \textbf{M}$\rightarrow$\textbf{U}, and \textbf{U}$\rightarrow$\textbf{M}.

\textbf{VisDA-2017}~\cite{visda2017} is a challenging benchmark, with $12$ classes shared by two extremely distinct domains: \textbf{Synthetic}, which contains $152,397$ synthetic images by rendering 3D models from different angles and under different lighting conditions, and \textbf{Real}, which comprises $55,388$ natural images. We evaluate on \textbf{Synthetic}$\rightarrow$\textbf{Real}.

%We show several samples from each domain in the appendix.

\subsection{Settings and Implementation Details}
\label{SecSetups}

We follow standard evaluation protocols for unsupervised domain adaptation~\cite{dann,cdan,sbada_gan}. We use all labeled source instances and all unlabeled target ones as training data, and report the average classification accuracy on target training data over three random trials. We use ResNet-50~\cite{resnet} pre-trained on ImageNet~\cite{imagenet} as the base network. Its last FC layer is replaced by two FC layers ($2048 \to 512 \to K$) and the loss terms of Fisher and ordering are minimized over both spaces $\cal{F}$ and ${\cal{F}}_{\uparrow}$. 
The lower convolutional and upper FC layers are used as the feature extractor $F$ and task classifier $C$ respectively. 
We follow~\cite{dann} to increase the hyper-parameter $\lambda$ from $0$ to $1$ by $\lambda_p=2(1+\exp(-\gamma p))^{-1}-1$, where $\gamma$ is set to $10$. We follow~\cite{mstn} to set the moving average coefficient $\alpha=0.7$. It is expected that the predicted confidence for any centroid should be higher than that for any instance, since correctly classifying the centroids is of more importance~\cite{ldada}. Thus, we empirically set the temperature $T=2$ for ordering losses and the second term of Fisher losses, and set $T=1$ for the remaining loss terms. 
We fine-tune $F$ and train $C$ from scratch, where the learning rate of $C$ is $10$ times that of $F$. We follow~\cite{dann} to employ the SGD training schedule: the learning rate of $C$ is adjusted by $\eta_p=\eta_0(1+\mu p)^{-\nu}$, where $p$ denotes the training epochs normalized to be in $[0,1]$, and we set $\eta_0=0.01$, $\mu=10$, and $\nu=0.75$. The momentum, weight decay, batch size, number of training epochs are set to $0.9$, $0.0001$, $64$, and $200$ respectively. 
For Digits, given that each domain has been split into the training and test sets, we follow~\cite{sbada_gan,mcd} and use all labeled instances from the source training set and all unlabeled ones from the target training set as training data. We report the average accuracy on the target test set over five random trials and adopt the same network architecture (i.e. LeNet~\cite{lenet}) and experimental setting as~\cite{sbada_gan,mcd}. 
For VisDA-2017, we follow~\cite{mcd} and report per-category and mean classification accuracy on $12$ classes; we use ImageNet pre-trained ResNet-101 as the base network; we set the initial learning rate $\eta_0$ and number of training epochs as $0.001$ and $20$ respectively. 
%Our implementation is based on PyTorch~\cite{pytorch} and 
The code is available at \url{https://github.com/huitangtang/DisClusterDA}. 

\begin{table*}[!t]
	\centering
	%\footnotesize
	\caption{Ablation study. Please refer to the main text for how these methods are defined.}
	\label{table:results_ablation_study} 
	\resizebox{0.9\textwidth}{!}{
		\begin{tabular}{lccccccc}
			\hline
			Methods                & A$\rightarrow$D & D$\rightarrow$A & Ar$\rightarrow$Pr & Pr$\rightarrow$Ar & M$\rightarrow$U & U$\rightarrow$M & Avg \\
			\hline
			Source Only                                   & 82.1 & 64.5 & 67.2 & 54.7 & 68.1 & 79.5 & 69.4 \\
			
			DisClusterDA (replacing afem with em)         & 94.7 & 75.5 & 74.7 & 63.3 & 92.9 & 95.4 & 82.8 \\ 
			
			DisClusterDA (w/o Fisher and ordering)        & 91.4 & 69.0 & 70.5 & 56.0 & 93.7 & 86.5 & 77.9 \\
			
			DisClusterDA (w/o Fisher)                     & 95.0 & 75.5 & 76.2 & 64.6 & 94.6 & 95.8 & 83.6 \\
			
			DisClusterDA (w/o distilling)                 & 93.5 & 75.4 & 75.6 & 64.3 & 93.4 & 95.3 & 82.9 \\
			
			DisClusterDA (w/o source ordering)            & 94.7 & 75.4 & 76.3 & 62.4 & 94.4 & 95.7 & 83.2 \\
			
			DisClusterDA (w/o source Fisher)              & 94.9 & 75.4 & 76.5 & 65.1 & 95.2 & 96.3 & 83.9 \\
			
			DisClusterDA (w/o temperature)                & 95.1 & 75.0 & 75.7 & 63.2 & 93.2 & 95.9 & 83.0 \\ 
			
			DisClusterDA (adding explicit domain alignment)             & 96.5 & 75.4 & 76.1 & 64.4 & 93.7 & 95.9 & 83.7 \\
			
			\hline
			\textbf{DisClusterDA}                                  & \textbf{96.8} & \textbf{76.5} & \textbf{77.0} & \textbf{65.9} & \textbf{95.6} & \textbf{96.6} & \textbf{84.7} \\
			\hline
		\end{tabular}
	}
\end{table*} 

\subsection{Ablation Study}

To empirically investigate the effects of components of DisClusterDA, we perform ablation study on six different adaptation tasks of \textbf{A}$\rightarrow$\textbf{D}, \textbf{D}$\rightarrow$\textbf{A}, \textbf{Ar}$\rightarrow$\textbf{Pr}, \textbf{Pr}$\rightarrow$\textbf{Ar}, \textbf{M}$\rightarrow$\textbf{U}, and \textbf{U}$\rightarrow$\textbf{M} by evaluating several variants of our method: 1) Source Only, which learns a standard classification network on labeled source data; 2) DisClusterDA (replacing afem with em), which replaces the proposed adaptive filtering entropy minimization loss with the original one; 3) DisClusterDA (w/o Fisher and ordering), which removes loss terms of Fisher and ordering from the overall objective (\ref{EqnOverallObj}); 4) DisClusterDA (w/o Fisher), which removes loss terms of Fisher; 5) DisClusterDA (w/o distilling), which removes the loss term of distilling and fine-tunes a trained Source Only model; 6) DisClusterDA (w/o source ordering), which removes the loss term of source ordering; 7) DisClusterDA (w/o source Fisher), which removes the loss term of source Fisher; 8) DisClusterDA (w/o temperature), which removes the temperature (i.e., $T=1$ in (\ref{EqnSoftMaxTemperature})); 9) DisClusterDA (adding explicit domain alignment), which adds a loss term of minimizing distances between corresponding source and target centroids. 

The results are reported in Table~\ref{table:results_ablation_study}. We can observe that the performance degrades when any one of our designed components is removed, verifying that all components of our DisClusterDA are complementary. 
Table~\ref{table:results_ablation_study} also tells that cluster ordering via centroid classification is the most important component, for which a geometric intuition is given in Fig.~\ref{fig:geo_exp}. 
It is further observed that the target ordering loss has a more significant impact on the model performance. The reasons are as follows. \textbf{1)} The goal of unsupervised domain adaptation is to correctly classify samples from the target domain. \textbf{2)} The $K$ target cluster centroids are also from the target domain and thus there does not exist the domain gap. \textbf{3)} They are also termed as prototypes \cite{tpn,prototypical_networks}, which best characterize the semantics of a specific target class. Thus, they are more important than a general target instance, as told by the recent work \cite{ldada}. Therefore, the target centroids will bring more benefits to the model performance than the source ones. 
DisClusterDA improves over DisClusterDA (w/o source Fisher) and DisClusterDA (w/o source ordering) on all adaptation tasks, highlighting the significance of enforcing similar cluster structures between the source and target domains. 
%We note that in Table \ref{table:results_ablation_study}, without the source ordering loss, the average performance degrades by $1.5\%$, which is considerable. 
Besides, DisClusterDA outperforms DisClusterDA (replacing afem with em), testifying the effectiveness of our adaptive filtering entropy minimization loss. 
Notably, DisClusterDA (adding explicit domain alignment) performs worse than DisClusterDA, corroborating our motivation.

\begin{table*}[!t]
	\centering
	\caption{Comparison with typical clustering algorithms on Office-31 based on ResNet-50.}
	\label{table:comp_clust_algs_office31}
	\resizebox{0.9\textwidth}{!}{
		\begin{tabular}{lccccccc}
			\hline
			Methods                & A$\rightarrow$W & D$\rightarrow$W & W$\rightarrow$D & A$\rightarrow$D & D$\rightarrow$A & W$\rightarrow$A & Avg \\
			\hline
			
			$K$-means & 87.3$\pm$0.6 & 97.3$\pm$0.2 & 99.6$\pm$0.2 & 87.0$\pm$0.9 & 72.8$\pm$0.3 & 75.9$\pm$0.2 & 86.7 \\
			
			Spherical $K$-means & 89.1$\pm$0.8 & 97.0$\pm$0.5 & 99.6$\pm$0.0 & 87.3$\pm$0.9 & 73.1$\pm$0.1 & 74.8$\pm$0.1 & 86.8 \\
			
			Kernel $K$-means~\cite{kernel_clust} & 88.7$\pm$0.1 & 97.4$\pm$0.0 & 99.6$\pm$0.0 & 85.6$\pm$0.1 & 73.2$\pm$0.0 & 74.5$\pm$0.2 & 86.5 \\
			
			EM~\cite{ssl_em} & 89.5$\pm$0.3 & 98.8$\pm$0.2 & \textbf{100.0}$\pm$0.0 & 89.2$\pm$0.4 & 71.2$\pm$0.1 & 67.5$\pm$0.4 & 86.0 \\
			
			DIRT-T~\cite{dirt_t} & 90.7$\pm$0.2 & 98.7$\pm$0.1 & \textbf{100.0}$\pm$0.0 & 90.2$\pm$0.2 & 73.1$\pm$1.0 & 67.9$\pm$0.5 & 86.8 \\
			
			DEPICT~\cite{depict} & 92.6$\pm$0.2 & \textbf{99.2}$\pm$0.0 & \textbf{100.0}$\pm$0.0 & 91.6$\pm$0.4 & 76.0$\pm$0.3 & 75.5$\pm$0.4 & 89.2 \\
			\hline
			\textbf{DisClusterDA}              & \textbf{95.2}$\pm$0.2 & \textbf{99.2}$\pm$0.1 & \textbf{100.0}$\pm$0.0 & \textbf{96.8}$\pm$0.5 & \textbf{76.5}$\pm$0.1 & \textbf{77.0}$\pm$0.1 & \textbf{90.8} \\
			
			\hline
		\end{tabular}
	}
\end{table*}

\subsection{Comparison with Typical Clustering Algorithms} 

We compare the proposed DisClusterDA with several typical clustering algorithms, i.e. $K$-means, Spherical $K$-means, Kernel $K$-means~\cite{kernel_clust}, EM~\cite{ssl_em}, DIRT-T~\cite{dirt_t}, and DEPICT~\cite{depict}. For $K$-means based clustering algorithms, we iteratively update pseudo labels for target data via cluster assignment and then train the same classification model $C \circ F$ with both pseudo-labeled target data and labeled source data. For EM, DIRT-T, and DEPICT, we follow their respective papers to cluster unlabeled target data while learning the model on labeled source data. Table~\ref{table:comp_clust_algs_office31} shows the results on Office-31 based on ResNet-50. It is observed that DisClusterDA significantly outperforms the compared clustering algorithms, confirming that DisClusterDA based on our proposed adaptive filtering entropy minimization, soft Fisher-like criterion, and cluster ordering via centroid classification can produce better clustering solutions.

\begin{figure}[!t]
	\centering
	\subfigcapskip=-8pt
	\subfigure[\textbf{A}$\rightarrow$\textbf{D}]{\includegraphics[height=1.3in]{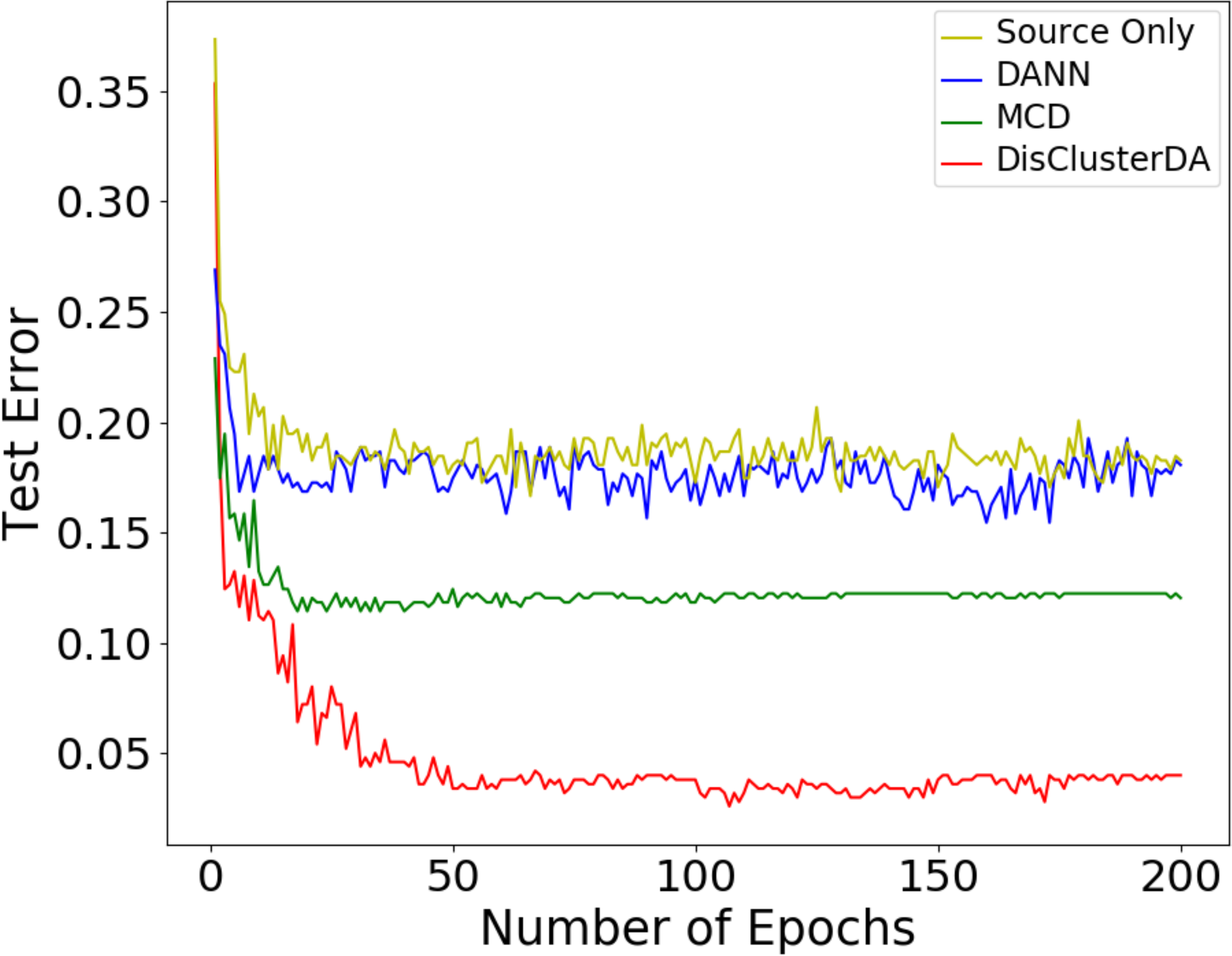}
		\label{fig:convergence1}}
	\hspace{0.2in}
	\subfigure[\textbf{D}$\rightarrow$\textbf{A}]{\includegraphics[height=1.3in]{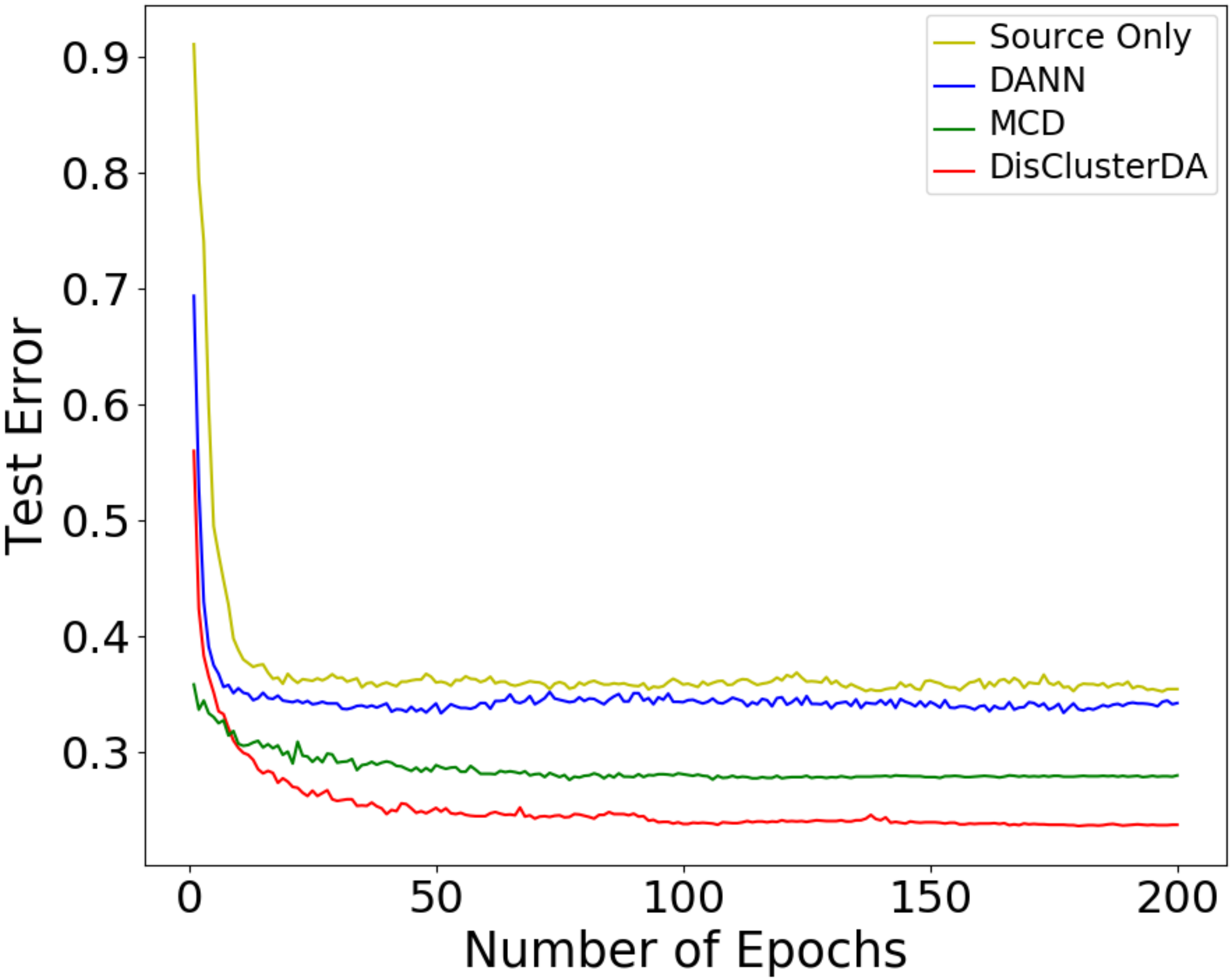}
		\label{fig:convergence2}}
	\vspace{-0.3cm}
	\caption{Convergence performance on the adaptation tasks of \textbf{A}$\rightarrow$\textbf{D} and \textbf{D}$\rightarrow$\textbf{A}. %Best viewed in color.
	}
	\label{fig:convergence}%\vspace{-0.2cm}
\end{figure}

\begin{figure}[!t]
	\centering
	\subfigbottomskip=0.1pt
	\subfigcapskip=-10pt
	\subfigure[Source Only]{\includegraphics[height=1.5in]{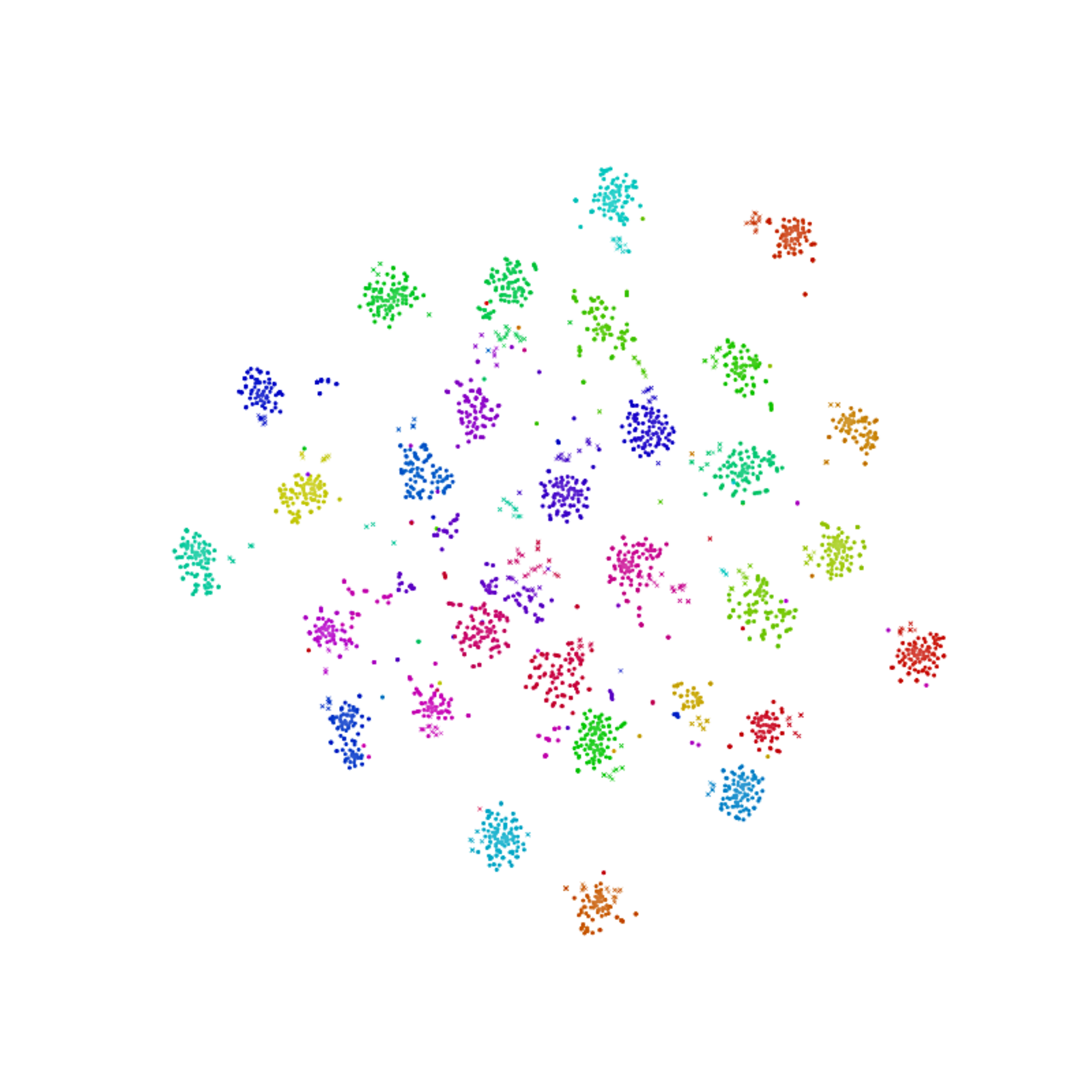}
		\label{fig:visualization1}}
	\hspace{0.5in}
	\subfigure[DANN]{\includegraphics[height=1.5in]{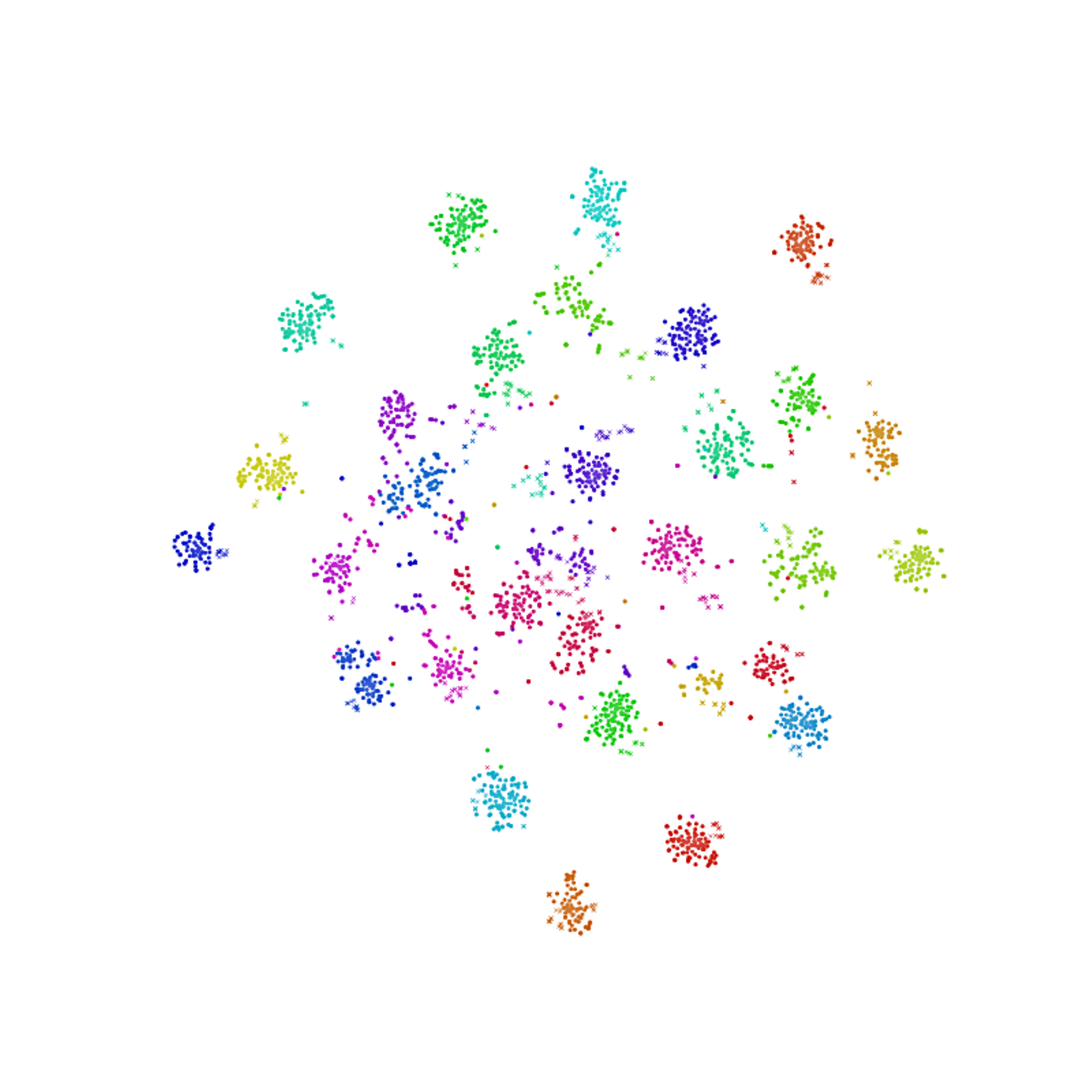}
		\label{fig:visualization2}}
	\\ \vspace{-0.13cm}
	\subfigure[MCD]{\includegraphics[height=1.5in]{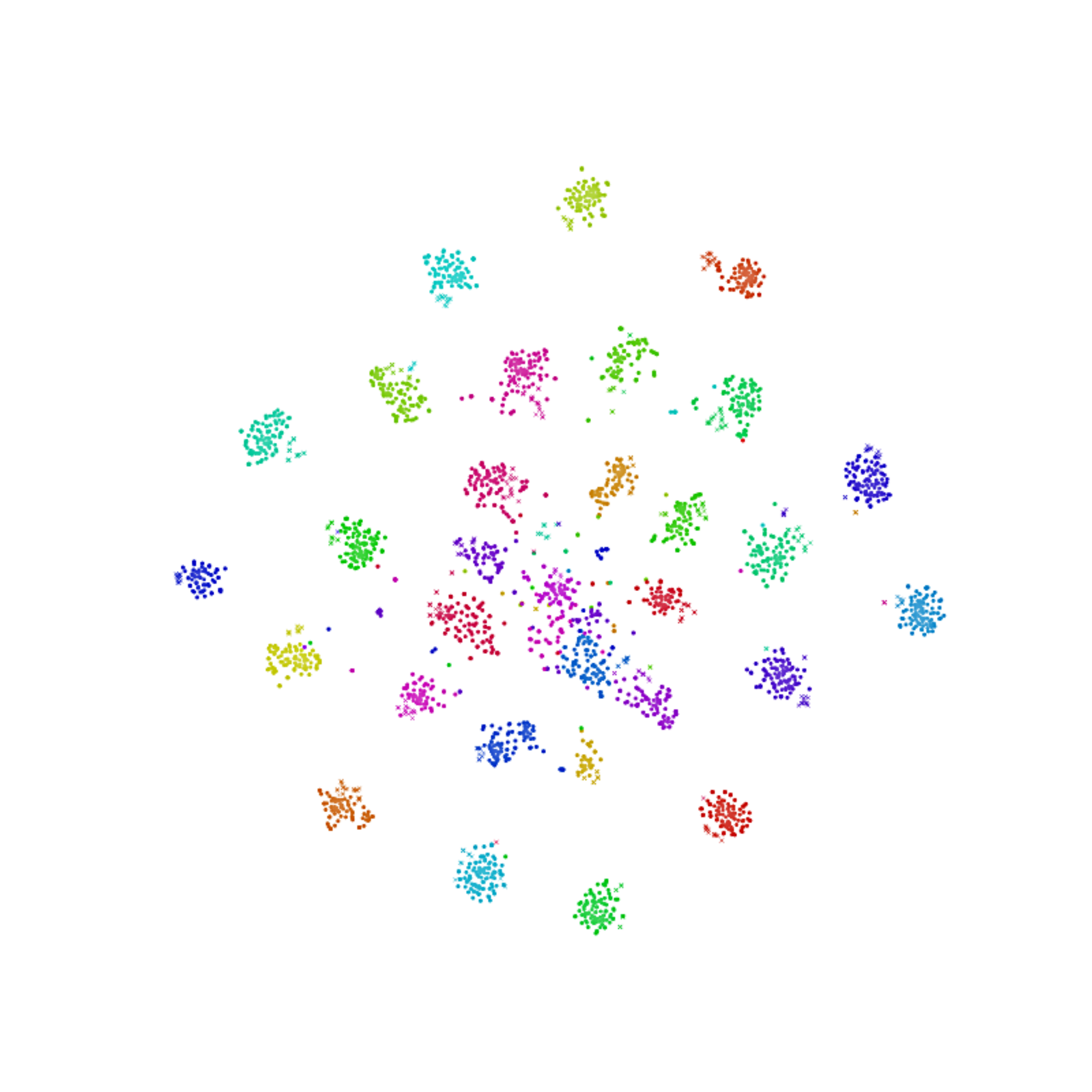}
		\label{fig:visualization3}}
	\hspace{0.5in}
	\subfigure[DisClusterDA]{\includegraphics[height=1.5in]{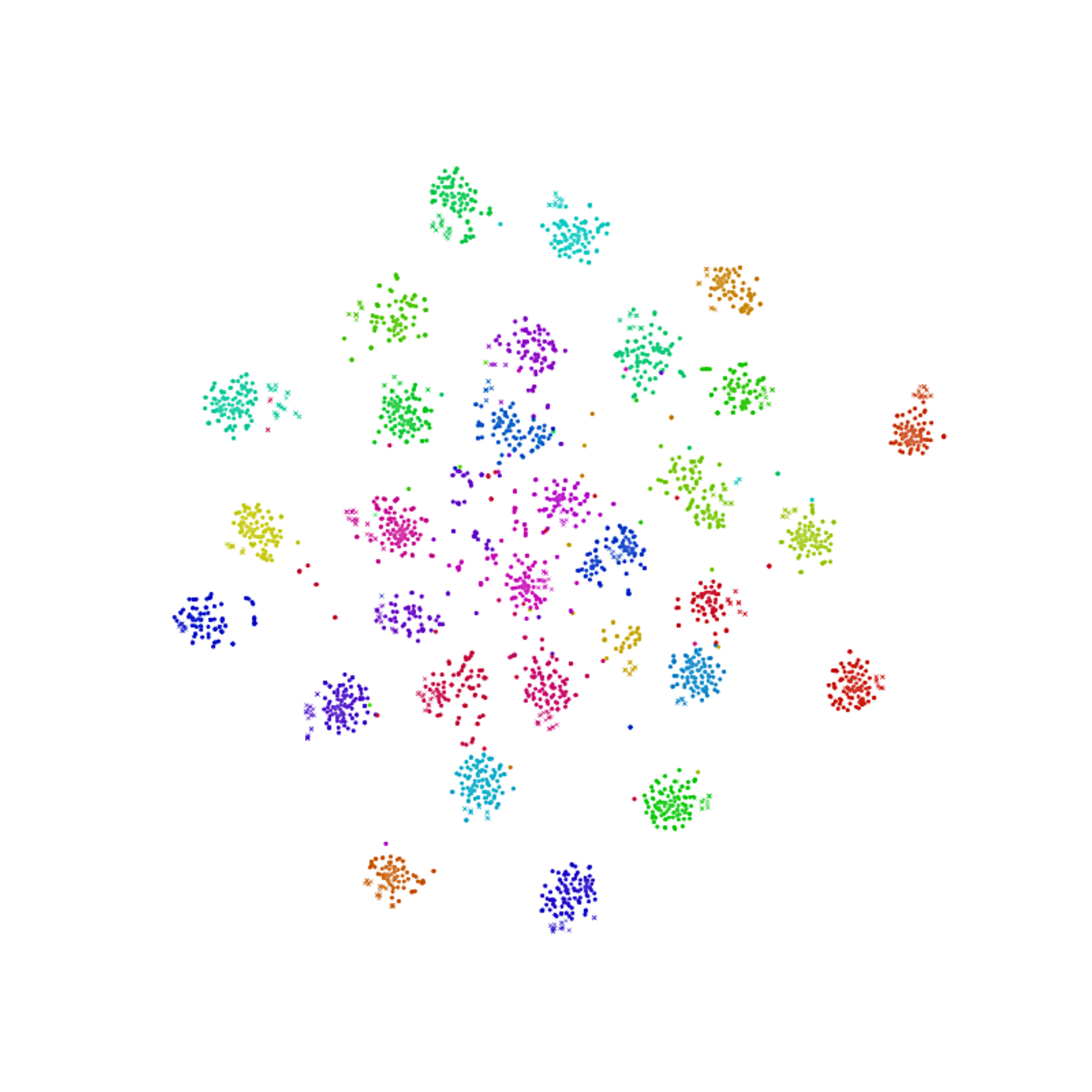}
		\label{fig:visualization4}}
	\\ \vspace{-0.13cm}
	\subfigure[Source Only]{\includegraphics[height=1.5in]{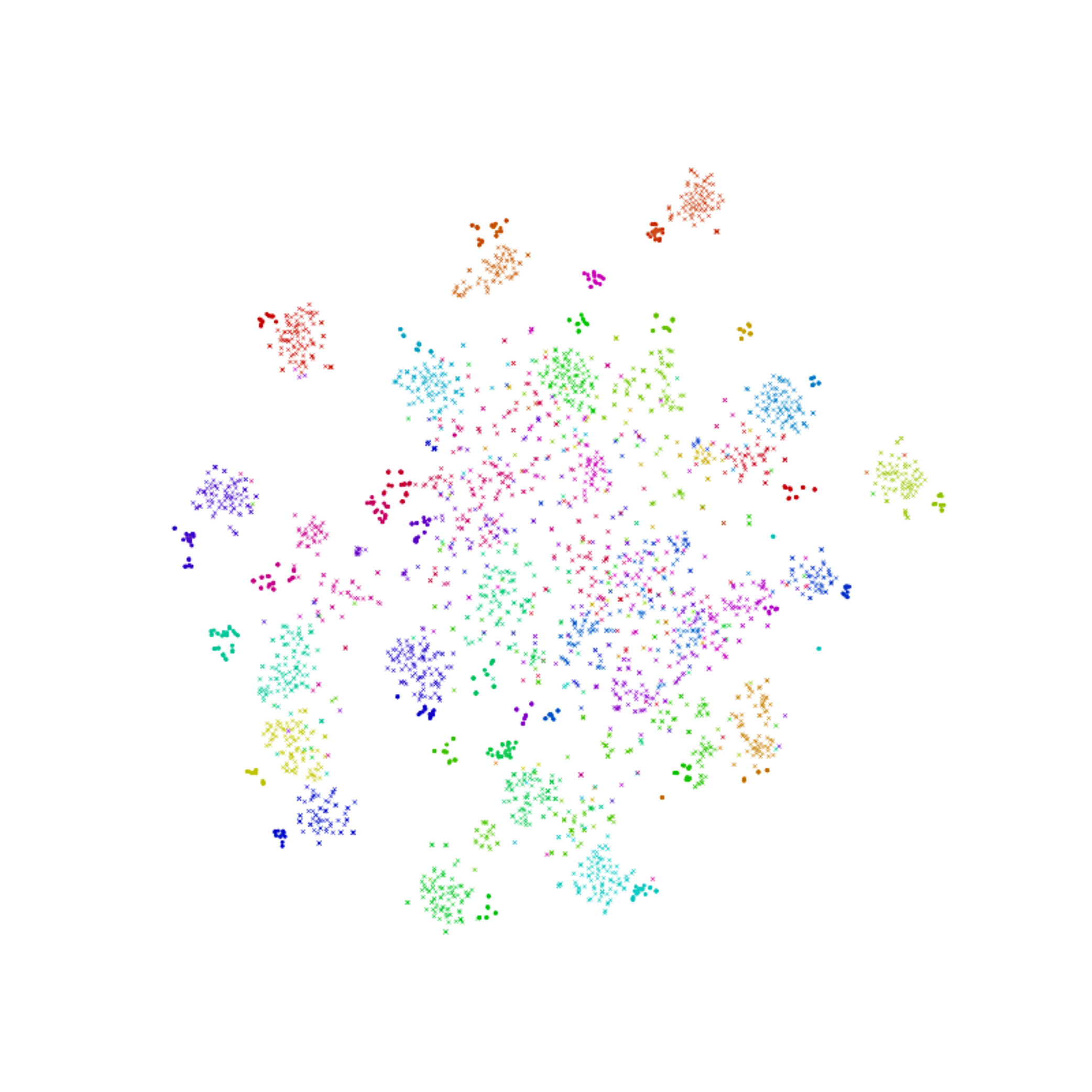}
		\label{fig:visualization5}}
	\hspace{0.5in}
	\subfigure[DANN]{\includegraphics[height=1.5in]{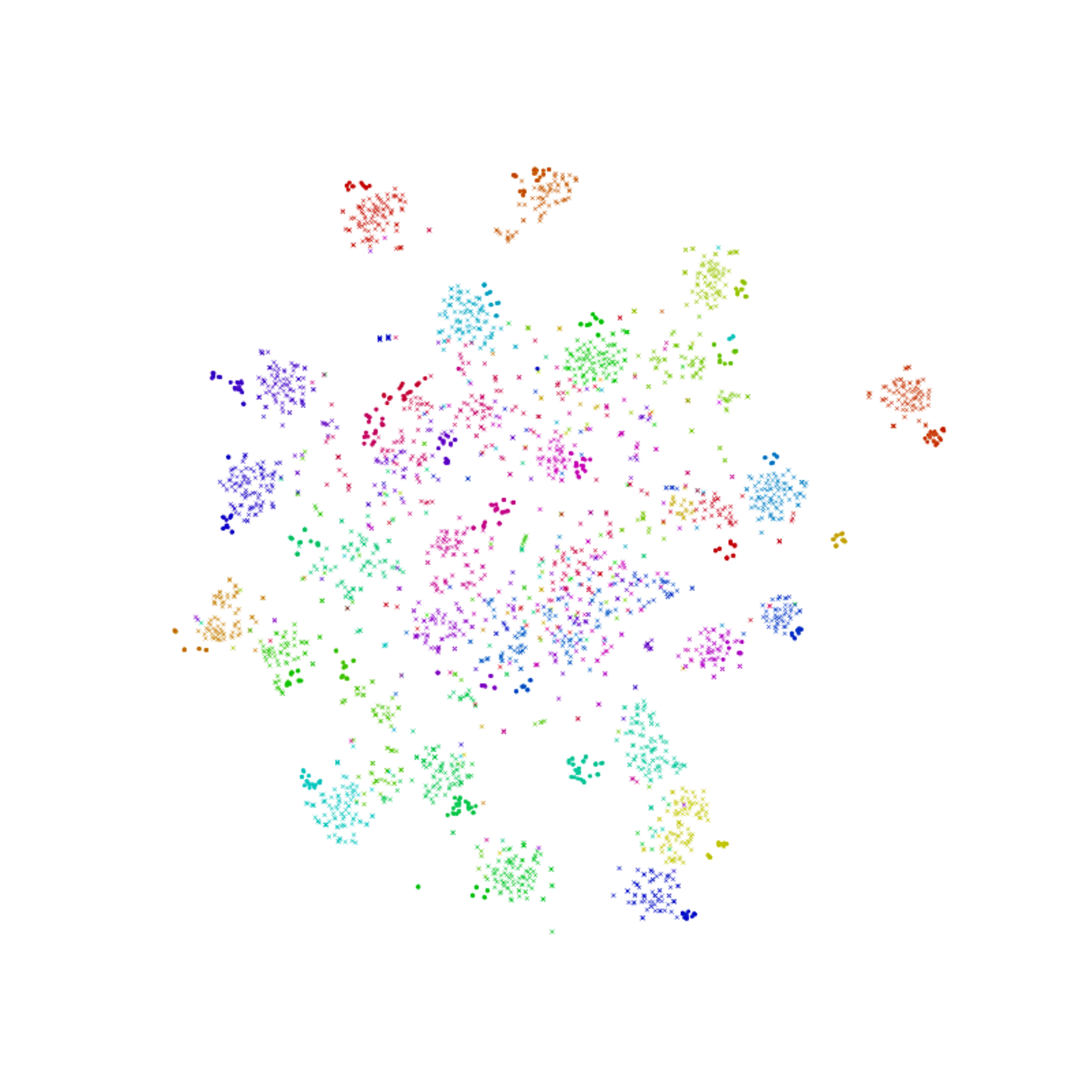}
		\label{fig:visualization6}}
	\\ \vspace{-0.13cm}
	\subfigure[MCD]{\includegraphics[height=1.5in]{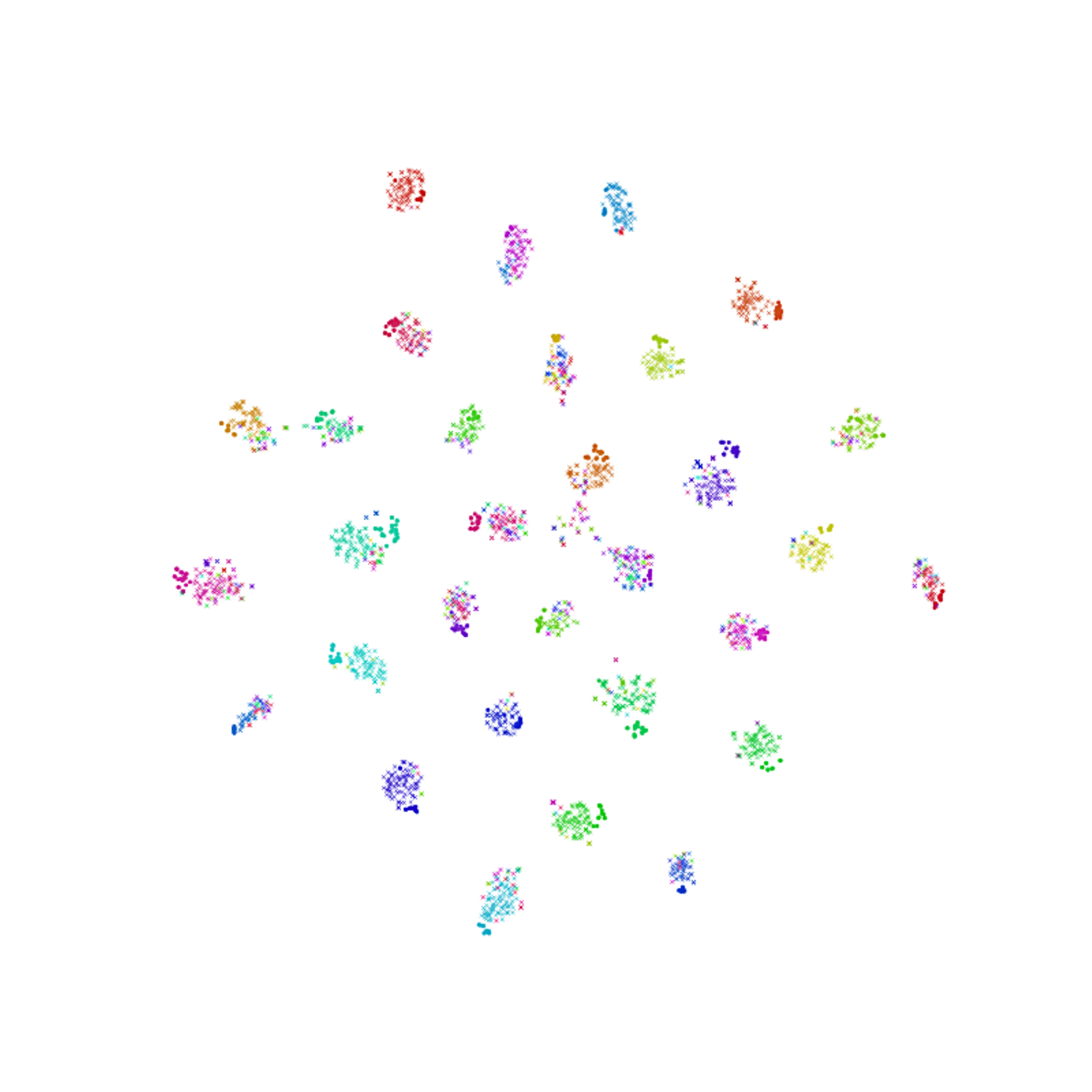}
		\label{fig:visualization7}}
	\hspace{0.5in}
	\subfigure[DisClusterDA]{\includegraphics[height=1.5in]{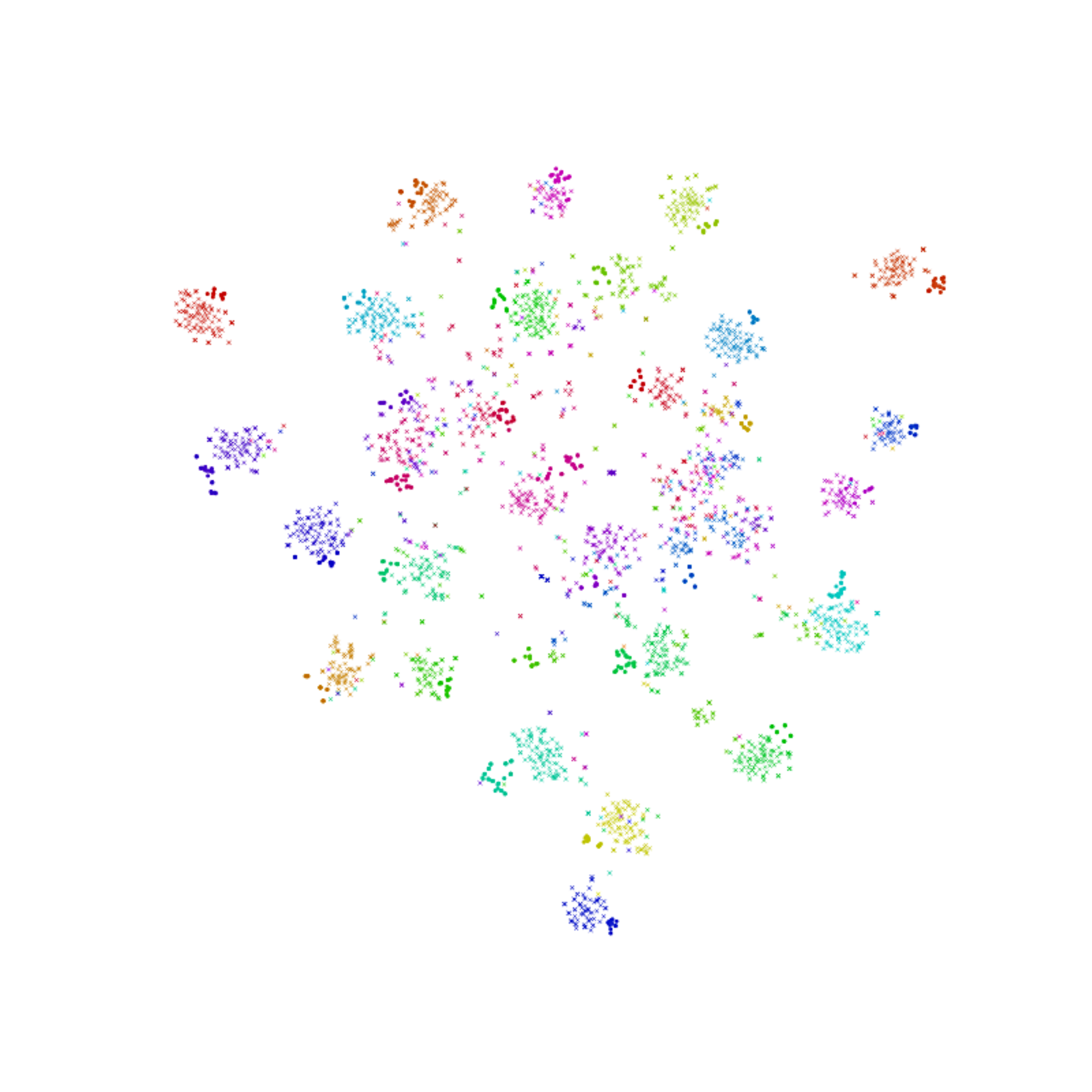}
		\label{fig:visualization8}}
	\vspace{-0.3cm}
	\caption{The t-SNE visualization of both domain features on the adaptation tasks of \textbf{A}$\rightarrow$\textbf{D} ((a)-(d)) and \textbf{D}$\rightarrow$\textbf{A} ((e)-(h)). Note that the markers ``o'' and ``x'' represent the source and target domains respectively, and different colors denote different classes. (Zoom in to see the details.) %Best viewed in color.
	}
	\label{fig:visualization}%\vspace{-0.2cm}
\end{figure}

\subsection{Convergence Performance and Feature Visualization}

In Fig.~\ref{fig:convergence}, we compare convergence performances of Source Only, DANN, MCD, and DisClusterDA in terms of test errors on the adaptation tasks of \textbf{A}$\rightarrow$\textbf{D} and \textbf{D}$\rightarrow$\textbf{A}. We can observe that the test error of each method first decreases quickly and then stabilizes at a certain level; notably, our proposed DisClusterDA consistently converges much better than the compared methods.

In Fig.~\ref{fig:visualization}, we use t-SNE~\cite{tsne} to visualize both domain features on the adaptation tasks of \textbf{A}$\rightarrow$\textbf{D} and \textbf{D}$\rightarrow$\textbf{A}, which are extracted by the feature extractors of Source Only, DANN~\cite{dann}, MCD~\cite{mcd}, and DisClusterDA. 
%It is observed that the feature alignment is better achieved by DANN and MCD. They enforce explicit domain-level and class-level feature alignments via domain-adversarial training and minimizing maximum classifier discrepancy respectively, which cause the undesired false alignment between different classes. Such a false alignment damages the diversity of the learned feature representation, which has negative effects on the model generalization~\cite{mcr2}. In contrast, with no explicit feature alignment, DisClusterDA preserves the feature diversity while making better achievements in the most important cluster discrimination, confirming the excellent effect of our proposed method on feature representation learning.
We observe that the adversarial training based methods, such as MCD \cite{mcd}, align features between the source and target domains with greater intensity than our proposed DisClusterDA, and thus the clusters seem to be a little more concentrated. However, such an explicitly enforced feature alignment has brought about possibly irreversible negative effects, i.e., \emph{the samples from different classes are catastrophically overlapped} (see the data point distribution in several clusters at the center of (c) and most clusters in (g)). As a result, both the intrinsic discriminative structures of target data and the feature diversity are damaged, which are adverse to the future model generalization~\cite{on_learn_invariant_rep,mcr2}. Compared to (c) and (g) of MCD, our proposed DisClusterDA of implicit domain alignment \emph{avoids the severe feature misalignment while preserving the feature diversity} (cf. (d) and (h)), leading to the greatly improved intra-cluster purity and diversity and inter-cluster discrimination.

\begin{table*}[!t]
	\centering
	%\footnotesize
	\caption{Sensitivity to moving average coefficient $\alpha$ on Office-31 based on ResNet-50.}
	\label{table:sen_alpha_office31} \vspace{-0.2cm}
	\resizebox{0.85\textwidth}{!}{
		\begin{tabular}{lccccccc}
			\hline
			Methods                & A$\rightarrow$W & D$\rightarrow$W & W$\rightarrow$D & A$\rightarrow$D & D$\rightarrow$A & W$\rightarrow$A & Avg \\
			\hline
			DisClusterDA ($\alpha=0.5$)             & 93.6$\pm$0.2 & 99.2$\pm$0.1 & \textbf{100.0}$\pm$0.0 & 94.4$\pm$0.5 & 75.1$\pm$0.4 & 75.0$\pm$0.3 & 89.6 \\
			
			DisClusterDA ($\alpha=0.6$)             & 94.2$\pm$0.2 & 99.2$\pm$0.1 & \textbf{100.0}$\pm$0.0 & 94.9$\pm$0.6 & 74.4$\pm$0.7 & 73.5$\pm$0.4 & 89.4 \\
			
			DisClusterDA ($\alpha=0.7$)             & \textbf{95.2}$\pm$0.2 & \textbf{99.2}$\pm$0.1 & \textbf{100.0}$\pm$0.0 & \textbf{96.8}$\pm$0.5 & \textbf{76.5}$\pm$0.1 & \textbf{77.0}$\pm$0.1 & \textbf{90.8} \\
			
			DisClusterDA ($\alpha=0.8$)             & 93.8$\pm$0.4 & 99.1$\pm$0.2 & \textbf{100.0}$\pm$0.0 & 93.8$\pm$0.4 & 73.2$\pm$0.4 & 73.6$\pm$0.8 & 88.9 \\
			
			DisClusterDA ($\alpha=0.9$)             & 94.0$\pm$0.7 & 99.0$\pm$0.2 & \textbf{100.0}$\pm$0.0 & 94.6$\pm$0.9 & 74.0$\pm$0.5 & 73.5$\pm$0.7 & 89.2 \\
			
			\hline
		\end{tabular} 
	}
\end{table*}%\vspace{-0.3cm}

\begin{table*}[!t]
	\centering
	%\footnotesize
	\caption{Sensitivity to temperature $T$ on Office-31 based on ResNet-50.}
	\label{table:sen_temperature_office31} \vspace{-0.2cm}
	\resizebox{0.85\textwidth}{!}{
		\begin{tabular}{lccccccc}
			\hline
			Methods                & A$\rightarrow$W & D$\rightarrow$W & W$\rightarrow$D & A$\rightarrow$D & D$\rightarrow$A & W$\rightarrow$A & Avg \\
			\hline
			DisClusterDA ($T=1$)             & 94.3$\pm$0.1 & 99.2$\pm$0.1 & \textbf{100.0}$\pm$0.0 & 95.1$\pm$0.1 & 75.0$\pm$0.2 & 73.1$\pm$0.3 & 89.5 \\
			
			DisClusterDA ($T=2$)             & \textbf{95.2}$\pm$0.2 & \textbf{99.2}$\pm$0.1 & \textbf{100.0}$\pm$0.0 & \textbf{96.8}$\pm$0.5 & \textbf{76.5}$\pm$0.1 & \textbf{77.0}$\pm$0.1 & \textbf{90.8} \\
			
			DisClusterDA ($T=3$)             & 93.2$\pm$0.4 & 99.2$\pm$0.1 & \textbf{100.0}$\pm$0.0 & 96.1$\pm$0.2 & 75.1$\pm$0.3 & 75.1$\pm$0.1 & 89.8 \\
			
			DisClusterDA ($T=4$)             & 93.5$\pm$0.6 & 99.2$\pm$0.1 & \textbf{100.0}$\pm$0.0 & 94.2$\pm$0.9 & 75.2$\pm$0.3 & 74.7$\pm$0.3 & 89.5 \\
			
			DisClusterDA ($T=5$)             & 93.7$\pm$0.4 & 99.1$\pm$0.2 & \textbf{100.0}$\pm$0.0 & 93.8$\pm$0.3 & 76.1$\pm$0.4 & 74.7$\pm$0.5 & 89.6 \\
			
			\hline
		\end{tabular} 
	}
\end{table*}%\vspace{-0.3cm}

\begin{figure*}[!t]
	\centering
	\subfigcapskip=-8pt
	\subfigure[Sensitivity to $\alpha$]{\includegraphics[height=1.6in]{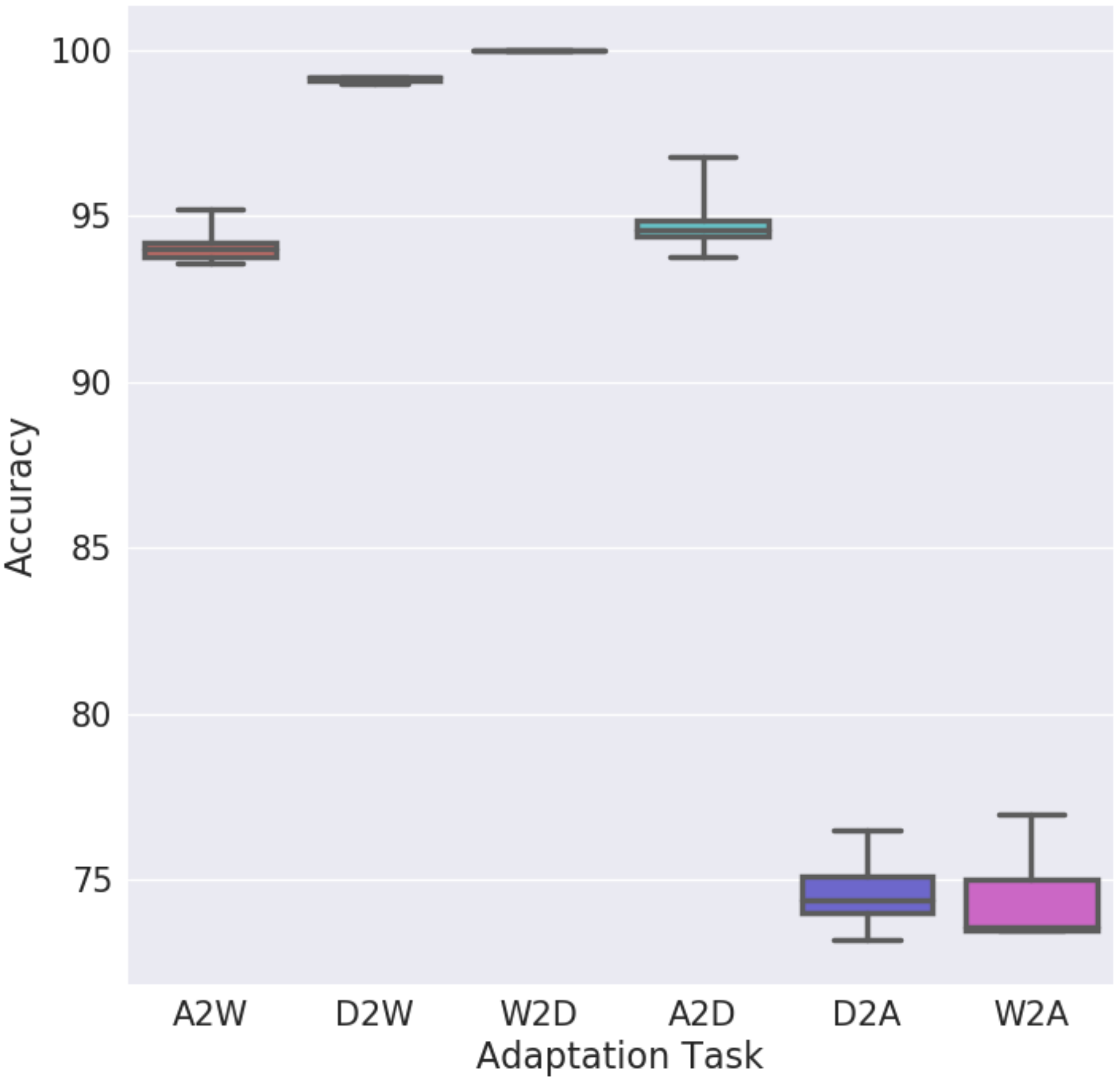}
		\label{fig:sensitivity1}}
	\hspace{0.2in}
	\subfigure[Sensitivity to $T$]{\includegraphics[height=1.6in]{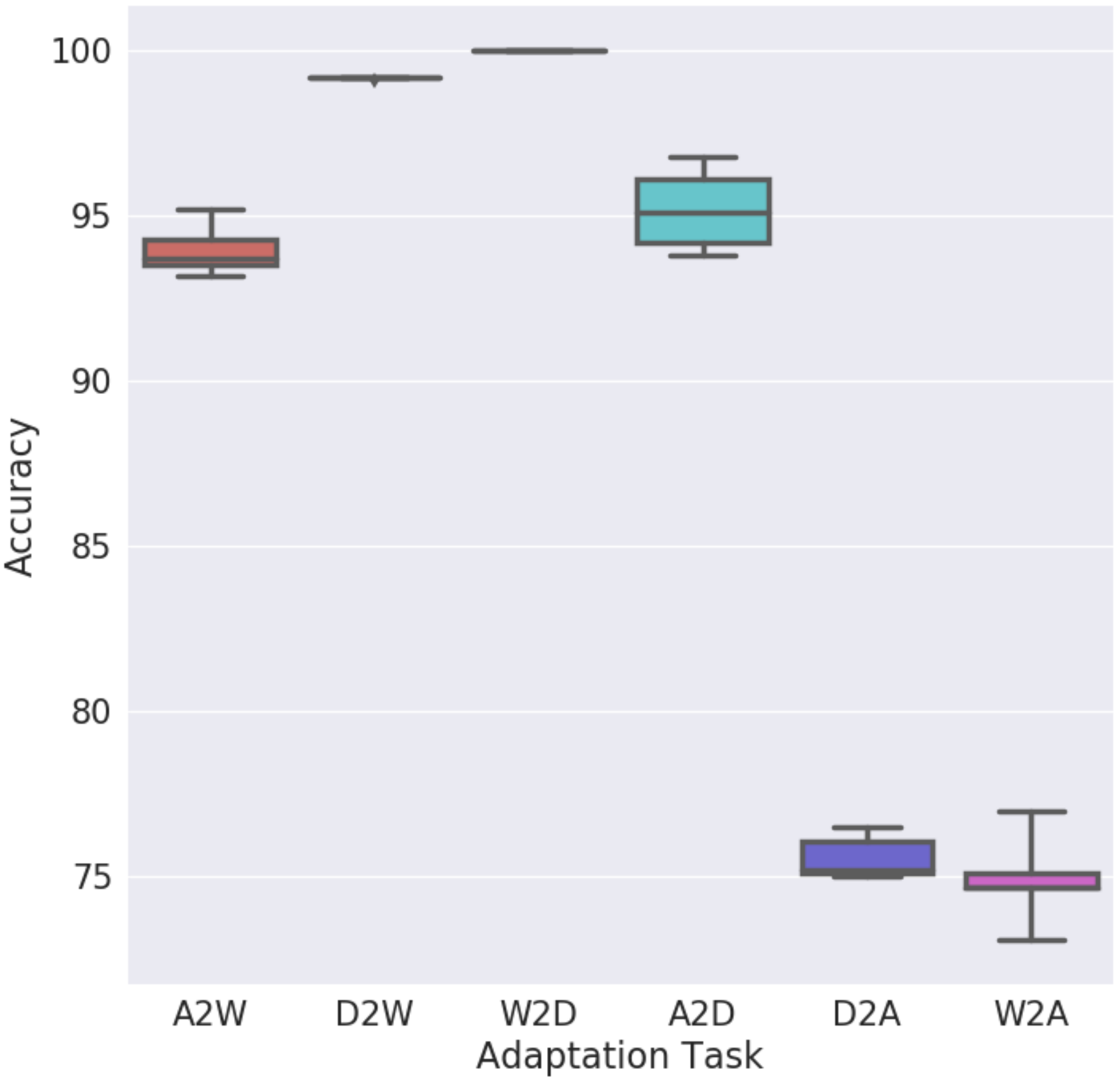}
		\label{fig:sensitivity2}}
	\vspace{-0.3cm}
	\caption{Sensitivity of DisClusterDA to the moving average coefficient $\alpha$ and temperature $T$. %Best viewed in color.
	}
	\label{fig:sensitivity}%\vspace{-0.2cm}
\end{figure*}

\subsection{Parameter Sensitivity and Efficiency Analysis}

We investigate the sensitivity of DisClusterDA to the moving average coefficient $\alpha$ and temperature $T$, by varying $\alpha \in \{0.5, 0.6, 0.7, 0.8, 0.9\}$ and $T \in \{1, 2, 3, 4, 5\}$ on Office-31 based on ResNet-50 in Tables~\ref{table:sen_alpha_office31} and \ref{table:sen_temperature_office31} respectively. The results of different variants on individual adaptation tasks are illustrated in Fig. \ref{fig:sensitivity}. We can observe that the accuracy fluctuation is very small when varying the value of $\alpha$ and $T$. Compared to $T$, DisClusterDA is more sensitive to the change of $\alpha$ in terms of the averaged accuracy. It is reasonable since the hyper-parameter $\alpha$ used in (\ref{EqnTargetCentroidMA}) is directly related to the reliability of source and target class centroids. In a vast range of $\alpha$ and $T$, DisClusterDA consistently outperforms the representative methods of DANN~\cite{dann} and MCD~\cite{mcd} (cf. Table~\ref{table:results_office31}), showing that DisClusterDA works reasonably stable.

Except for the time spent on forward/backward propagations of network training, which is common to existing deep unsupervised domain adaptation methods, the proposed method has an additional cost in the computation of cluster centroids (similar to~\cite{mstn}). This cost is, however, negligible since the number of clusters is the same as that of task categories, and clustering of any instance $\mathbf{x}$ is achieved on the fly from network prediction via $\arg\max_k p_k(\mathbf{x}), k \in {1, \dots, K}$.

\begin{table*}[!t]
	%\footnotesize
	\centering
	\caption{Results (\%) on Office-31 based on ResNet-50.}
	\label{table:results_office31}\vspace{-0.1cm}
	\resizebox{0.8\textwidth}{!}{
		\begin{tabular}{lccccccc}
			\hline
			Methods                & A$\rightarrow$W & D$\rightarrow$W & W$\rightarrow$D & A$\rightarrow$D & D$\rightarrow$A & W$\rightarrow$A & Avg \\
			\hline
			Source Only~\cite{resnet}           & 68.4$\pm$0.2 & 96.7$\pm$0.1 & 99.3$\pm$0.1 & 68.9$\pm$0.2 & 62.5$\pm$0.3 & 60.7$\pm$0.3 & 76.1 \\
			
			DANN~\cite{dann}                & 82.0$\pm$0.4 & 96.9$\pm$0.2 & 99.1$\pm$0.1 & 79.7$\pm$0.4 & 68.2$\pm$0.4 & 67.4$\pm$0.5 & 82.2 \\
			
			DAN~\cite{dan}                    & 86.3$\pm$0.3 & 97.2$\pm$0.2 & 99.6$\pm$0.1 & 82.1$\pm$0.3 & 64.6$\pm$0.4 & 65.2$\pm$0.3 & 82.5 \\
			
			JAN-A~\cite{jan}                   & 86.0$\pm$0.4 & 96.7$\pm$0.3 & 99.7$\pm$0.1 & 85.1$\pm$0.4 & 69.2$\pm$0.4 & 70.7$\pm$0.5 & 84.6 \\
			
			%VADA~\cite{dirt_t}              & 86.5$\pm$0.5 & 98.2$\pm$0.4 & 99.7$\pm$0.2 & 86.7$\pm$0.4 & 70.1$\pm$0.4 & 70.5$\pm$0.4 & 85.4 \\
			
			%GAACN~\cite{gaacn}             & 90.2 & 98.4 & \textbf{100.0} & 90.4 & 67.4 & 67.7 & 85.6 \\
			
			%ETD~\cite{etd}          & 92.1 & \textbf{100.0} & \textbf{100.0} & 88.0 & 71.0 & 67.8 & 86.2 \\
			
			MCD~\cite{mcd}                  & 88.6$\pm$0.2 & 98.5$\pm$0.1 & \textbf{100.0}$\pm$0.0 & 92.2$\pm$0.2 & 69.5$\pm$0.1 & 69.7$\pm$0.3 & 86.5 \\
			
			SAFN+ENT~\cite{larger_norm}   & 90.1$\pm$0.8 & 98.6$\pm$0.2 & 99.8$\pm$0.0 & 90.7$\pm$0.5 & 73.0$\pm$0.2 & 70.2$\pm$0.3 & 87.1 \\
			
			rRevGrad+CAT~\cite{cat}   & 94.4$\pm$0.1 & 98.0$\pm$0.2 & \textbf{100.0}$\pm$0.0 & 90.8$\pm$1.8 & 72.2$\pm$0.6 & 70.2$\pm$0.1 & 87.6 \\ 
			
			CTSN~\cite{ctsn}              & 90.6$\pm$0.3 & 98.6$\pm$0.5 & 99.9$\pm$0.1 & 89.3$\pm$0.3 & 73.7$\pm$0.4 & 74.1$\pm$0.3 & 87.7 \\
			
			DMRL~\cite{dmrl}  & 90.8$\pm$0.3 & 99.0$\pm$0.2 & \textbf{100.0}$\pm$0.0 & 93.4$\pm$0.5 & 73.0$\pm$0.3 & 71.2$\pm$0.3 & 87.9 \\
			
			MSTN+DSBN~\cite{dsbn}              & 92.7 & 99.0 & \textbf{100.0} & 92.2 & 71.7 & 74.4 & 88.3 \\%$\star$ 
			
			%SymNet~\cite{symnet}     & 90.8$\pm$0.1 & 98.8$\pm$0.3 & \textbf{100.0}$\pm$0.0 & 93.9$\pm$0.5 & 74.6$\pm$0.6 & 72.5$\pm$0.5 & 88.4 \\
			
			TAT~\cite{tat}                  & 92.5$\pm$0.3 & \textbf{99.3}$\pm$0.1 & \textbf{100.0}$\pm$0.0 & 93.2$\pm$0.2 & 73.1$\pm$0.3 & 72.1$\pm$0.3 & 88.4 \\
			
			BSP+CDAN~\cite{bsp}             & 93.3$\pm$0.2 & 98.2$\pm$0.2 & \textbf{100.0}$\pm$0.0 & 93.0$\pm$0.2 & 73.6$\pm$0.3 & 72.6$\pm$0.3 & 88.5 \\
			
			CDAN+BNM~\cite{bnm}             & 92.8 & 98.8 & \textbf{100.0} & 92.9 & 73.5 & 73.8 & 88.6 \\
			
			MDD~\cite{mdd}             & 94.5$\pm$0.3 & 98.4$\pm$0.1 & \textbf{100.0}$\pm$0.0 & 93.5$\pm$0.2 & 74.6$\pm$0.3 & 72.2$\pm$0.1 & 88.9 \\
			
			ViCatDA~\cite{vicatda} & 94.5$\pm$0.2 & 99.2$\pm$0.1 & \textbf{100.0}$\pm$0.0 & 92.3$\pm$0.1 & \textbf{76.5}$\pm$0.2 & 74.2$\pm$0.1 & 89.5 \\
			
			GSDA~\cite{gsda} & \textbf{95.7} & 99.1 & \textbf{100.0} & 94.8 & 73.5 & 74.9 & 89.7 \\
			
			\hline
			\textbf{DisClusterDA}              & 95.2$\pm$0.2 & 99.2$\pm$0.1 & \textbf{100.0}$\pm$0.0 & \textbf{96.8}$\pm$0.5 & \textbf{76.5}$\pm$0.1 & \textbf{77.0}$\pm$0.1 & \textbf{90.8} \\
			\hline
		\end{tabular}%\vspace{-0.2cm}
	}
\end{table*}

\begin{table*}[!t]
	\centering
	%\footnotesize
	\caption{Results (\%) on Office-Home based on ResNet-50.}
	\label{table:results_officehome}\vspace{-0.2cm}
	\resizebox{1.0\textwidth}{!}{
		\begin{tabular}{lccccccccccccc}
			\hline
			Methods                         & Ar$\rightarrow$Cl & Ar$\rightarrow$Pr & Ar$\rightarrow$Rw & Cl$\rightarrow$Ar & Cl$\rightarrow$Pr & Cl$\rightarrow$Rw & Pr$\rightarrow$Ar & Pr$\rightarrow$Cl & Pr$\rightarrow$Rw & Rw$\rightarrow$Ar & Rw$\rightarrow$Cl & Rw$\rightarrow$Pr    & Avg  \\
			\hline
			Source Only~\cite{resnet}     & 34.9      & 50.0     & 58.0      & 37.4      & 41.9      & 46.2     & 38.5     & 31.2     & 60.4     & 53.9     & 41.2     & 59.9 & 46.1 \\
			
			DAN~\cite{dan}                  & 43.6     & 57.0     & 67.9      & 45.8      & 56.5      & 60.4     & 44.0     & 43.6     & 67.7     & 63.1     & 51.5     & 74.3  & 56.3 \\
			
			DANN~\cite{dann}                & 45.6     & 59.3     & 70.1      & 47.0      & 58.5      & 60.9     & 46.1     & 43.7     & 68.5     & 63.2     & 51.8      & 76.8 & 57.6 \\
			
			JAN~\cite{jan}                  & 45.9     & 61.2     & 68.9      & 50.4      & 59.7      & 61.0     & 45.8     & 43.4     & 70.3     & 63.9     & 52.4      & 76.8 & 58.3 \\
			
			%SE~\cite{se}  & 48.8 & 61.8 & 72.8 & 54.1 & 63.2 & 65.1 & 50.6 & 49.2 & 72.3 & 66.1 & 55.9 & 78.7 & 61.5 \\
			
			DWT-MEC~\cite{dwt_mec} & 50.3 & 72.1 & 77.0 & 59.6 & 69.3 & 70.2 & 58.3 & 48.1 & 77.3 & 69.3 & 53.6 & 82.0 & 65.6 \\
			
			TAT~\cite{tat} & 51.6 & 69.5 & 75.4 & 59.4 & 69.5 & 68.6 & 59.5 & 50.5 & 76.8 & 70.9 & 56.6 & 81.6 & 65.8 \\
			
			%GAACN~\cite{gaacn}              & 53.1 & 71.5 & 74.6 & 59.9 & 64.6 & 67.0 & 59.2 & 53.8 & 75.1 & 70.1 & 59.3 & 80.9 & 65.8 \\
			
			BSP+CDAN~\cite{bsp} & 52.0 & 68.6 & 76.1 & 58.0 & 70.3 & 70.2 & 58.6 & 50.2 & 77.6 & 72.2 & 59.3 & 81.9 & 66.3 \\
			
			SAFN~\cite{larger_norm} & 52.0 & 71.7 & 76.3 & 64.2 & 69.9 & 71.9 & 63.7 & 51.4 & 77.1 & 70.9 & 57.1 & 81.5 & 67.3 \\
			
			%ETD~\cite{etd}          & 51.3 & 71.9 & \textbf{85.7} & 57.6 & 69.2 & 73.7 & 57.8 & 51.2 & 79.3 & 70.2 & 57.5 & 82.1 & 67.3 \\
			
			%SymNet~\cite{symnet} & 47.7 & 72.9 & 78.5 & 64.2 & 71.3 & 74.2 & 64.2 & 48.8 & 79.5 & \textbf{74.5} & 52.6 & 82.7 & 67.6 \\
			
			MDD~\cite{mdd}         & 54.9 & 73.7 & 77.8 & 60.0 & 71.4 & 71.8 & 61.2 & 53.6 & 78.1 & 72.5 & 60.2 & 82.3 & 68.1 \\
			
			ViCatDA~\cite{vicatda} & 50.9 & 74.7 & 78.8 & 64.8 & 71.7 & 74.4 & 64.5 & 52.4 & 80.4 & \textbf{74.5} & 57.4 & 83.2 & 69.0 \\
			
			CDAN+BNM~\cite{bnm}    & 56.2 & 73.7 & 79.0 & 63.1 & 73.6 & 74.0 & 62.4 & 54.8 & 80.7 & 72.4 & 58.9 & 83.5 & 69.4 \\
			
			GSDA~\cite{gsda} & \textbf{61.3} & 76.1 & 79.4 & 65.4 & 73.3 & 74.3 & 65.0 & 53.2 & 80.0 & 72.2 & \textbf{60.6} & 83.1 & 70.3 \\
			
			\hline
			\textbf{DisClusterDA} & 58.8 & \textbf{77.0} & \textbf{80.8} & \textbf{67.0} & \textbf{74.6} & \textbf{77.1} & \textbf{65.9} & \textbf{56.3} &\textbf{81.4} & 74.2 & 60.5 & \textbf{83.6} & \textbf{71.4}  \\
			\hline
		\end{tabular}
	}%\vspace{-0.2cm}
\end{table*}

\begin{table}[!t]
	\centering
	%\footnotesize
	\caption{Results (\%) on Digits based on LeNet.}
	\label{table:results_digits}\vspace{-0.2cm}
	\resizebox{0.6\linewidth}{!}{
	\begin{tabular}{lccccc}
		\hline
		Methods                 & M$\rightarrow$S & S$\rightarrow$M & M$\rightarrow$U & U$\rightarrow$M & Avg \\
		\hline
		Source Only~\cite{lenet}  & 26.0  & 60.1 & 78.9 & 57.1 & 55.5 \\
		
		DAN~\cite{dan}          & - & 73.5 & 80.3 & 77.8 & - \\
		
		DANN~\cite{dann}        & 35.7 & 73.9 & 85.1 & 73.0 & 66.9 \\
		
		DRCN~\cite{drcn}        & 40.1 & 82.0 & 91.8 & 73.7 & 71.9 \\
		
		%CoGAN~\cite{cogan}      & - & - & 91.2 & 89.1 & - \\
		
		ATDA~\cite{atda} & 52.8 & 86.2 & - & - & - \\
		
		CyCADA~\cite{cycada} & - & 90.4 & 95.6 & 96.5 & - \\
		
		%MSTN~\cite{mstn}        & - & 91.7 & 92.9 & - & - \\
		
		TPN~\cite{tpn}          & - & 93.0 & 92.1 & 94.1 & - \\
		
		%VADA~\cite{dirt_t}     & 47.5 & 97.9 & - & - & - \\
		
		%PFAN~\cite{pfan}        & 57.6 & 93.9 & 95.0 & - & - \\
		
		SBADA-GAN~\cite{sbada_gan} & \textbf{61.1} & 76.1 & \textbf{97.6} & 95.0 & 82.5 \\
		
		%ADR~\cite{adr}          & - & 94.1 & 91.3 & 91.5 & - \\
		
		%GAACN~\cite{gaacn}      & - & 94.6 & 95.4 & 98.3 & - \\
		
		%DM-ADA~\cite{dm_ada}    & - & 95.5 & 94.8 & 94.2 & - \\
		
		MCD~\cite{mcd}          & - & 96.2 & 94.2 & 94.1 & - \\
		
		DMRL~\cite{dmrl} & - & 96.2 & 96.1 & \textbf{99.0} & - \\
		
		ViCatDA~\cite{vicatda}  & - & 97.1 & 96.0 & 96.7 & - \\
		
		CTSN~\cite{ctsn}        & - & 97.1 & 96.1 & 97.3 & - \\ 
		
		\hline
		\textbf{DisClusterDA}    & 60.2 & \textbf{98.7} & 95.6 & 96.6 & \textbf{87.8} \\
		\hline
	\end{tabular}%\vspace{-0.1cm}
	}
\end{table}

\begin{table*}[!t]
	\centering
	%\footnotesize
	\caption{Results (\%) on VisDA-2017 based on ResNet-101.
	}
	\label{table:results_visda2017} \vspace{-0.2cm}
	\resizebox{1.0\textwidth}{!}{
		\begin{tabular}{lccccccccccccc}
			\hline
			Methods                & plane & bcycl & bus & car & horse & knife & mcycl & person & plant & sktbrd & train & truck & \emph{mean} \\
			\hline
			Source Only~\cite{resnet}  & 55.1 & 53.3 & 61.9 & 59.1 & 80.6 & 17.9 & 79.7 & 31.2 & 81.0 & 26.5 & 73.5 & 8.5 & 52.4 \\
			
			DANN~\cite{dann}  & 81.9 & 77.7 & 82.8 & 44.3 & 81.2 & 29.5 & 65.1 & 28.6 & 51.9 & 54.6 & 82.8 & 7.8 & 57.4 \\
			
			DAN~\cite{dan}         & 87.1 & 63.0 & 76.5 & 42.0 & 90.3 & 42.9 & 85.9 & 53.1 & 49.7 & 36.3 & 85.8 & 20.7 & 61.1 \\
			
			MCD~\cite{mcd} & 87.0 & 60.9 & 83.7 & 64.0 & 88.9 & 79.6 & 84.7 & 76.9 & 88.6 & 40.3 & 83.0 & 25.8 & 71.9 \\
			
			%GPDA~\cite{gpda}  & 83.0 & 74.3 & 80.4 & 66.0 & 87.6 & 75.3 & 83.8 & 73.1 & 90.1 & 57.3 & 80.2 & 37.9 & 73.3 \\
			
			%ADR~\cite{adr}    & 87.8 & 79.5 & 83.7 & 65.3 & 92.3 & 61.8 & 88.9 & 73.2 & 87.8 & 60.0 & 85.5 & 32.3 & 74.8 \\
			
			BSP+CDAN~\cite{bsp} & 92.4 & 61.0 & 81.0 & 57.5 & 89.0 & 80.6 & 90.1 & 77.0 & 84.2 & 77.9 & 82.1 & 38.4 & 75.9 \\
			
			ViCatDA~\cite{vicatda} & 93.9 & 67.3 & 78.6 & 66.9 & 89.3 & 88.4 & 91.0 & 77.9 & 90.2 & 68.2 & \textbf{88.4} & 31.8 & 77.7 \\
			
			MSTN+DSBN~\cite{dsbn} & 94.7 & \textbf{86.7} & 76.0 & 72.0 & 95.2 & 75.1 & 87.9 & 81.3 & 91.1 & 68.9 & 88.3 & 45.5 & 80.2 \\ %$\star$
			
			TPN~\cite{tpn}    & 93.7 & 85.1 & 69.2 & 81.6 & 93.5 & 61.9 & 89.3 & 81.4 & 93.5 & 81.6 & 84.5 & 49.9 & 80.4 \\
			
			\hline
			\textbf{DisClusterDA} & \textbf{96.4} & 83.2 & \textbf{85.8} & \textbf{85.1} & \textbf{96.7} & \textbf{93.5} & \textbf{93.0} & \textbf{86.8} & \textbf{96.5} & \textbf{90.5} & 85.6 & \textbf{51.6} & \textbf{87.1} \\
			\hline
		\end{tabular}
	}
\end{table*}

\subsection{Experimental Results and Comparative Analyses}
\label{SecExpResults}

In this section, we compare the proposed DisClusterDA with state-of-the-art deep methods on four commonly used benchmark datasets of Office-31, Office-Home, Digits, and VisDA-2017 in Tables~\ref{table:results_office31}, \ref{table:results_officehome}, \ref{table:results_digits}, and \ref{table:results_visda2017} respectively, where results of the compared methods are quoted from their respective papers or~\cite{cdan,tat,sbada_gan,mcd,dwt_mec}. 
We highlight several interesting observations below. 
1) Although Source Only avoids distorting the intrinsic structures of target data, it performs worse due to a lack of knowledge transfer from the source domain to the target one. 
2) Explicit feature alignment methods (e.g. DANN and MCD), which could hurt the intrinsic target structures, exceed Source Only by a large margin, suggesting the importance of knowledge transfer. 
3) Class-level feature alignment methods (e.g. MCD and MDD), achieve much better results than domain-level ones (e.g. DANN and DAN), indicating the necessity of utilizing the semantic information of target data. 
4) DisClusterDA outperforms all compared methods and achieves the new state of the art on all these datasets, demonstrating the superiority of our proposed knowledge transfer method, which aims to preserve the intrinsic target discrimination; particularly, our method shows better generalization performance, given that the results on SVHN are measured on unseen instances sampled from the same target domain. 
5) DisClusterDA significantly improves the classification of target data on hard adaptation tasks, e.g., \textbf{A}$\rightarrow$\textbf{D} and \textbf{S}$\rightarrow$\textbf{M}, and on the difficult dataset of Office-Home, which still has a large room of improvement since it contains visually more dissimilar domains with more classes. 
6) DisClusterDA consistently remains superior whether the network is small (e.g. LeNet) or big (e.g. ResNet-101) and whether the domain is small-scale (e.g. Office-31) or large-scale (e.g. VisDA-2017). 
7) For the challenging yet realistically significant task \textbf{Synthetic}$\rightarrow$\textbf{Real}, DisClusterDA holds a remarkable gain over all compared methods on almost all object categories, especially those long-tailed ones with much fewer samples, e.g. knife and skateboard.

\subsection{Multi-Source Domain Adaptation}
\label{SecExpMSDA}

The conventional domain adaptation assumes a single source, i.e., the source instances are sampled from a single domain. However, in more practical scenarios where the labeled data are collected from multiple domains (e.g., with different camera and lighting conditions), the assumption could be violated, resulting in degraded performance for most of existing methods~\cite{dann,mcd}. To validate the generality and robustness of our proposed DisClusterDA, we do experiments for multi-source domain adaptation (MSDA) by combining multiple source domains as a single one. 
We use the widely used benchmark dataset \textbf{Office-Caltech10}~\cite{gfk}, which is extended from Office-31~\cite{office31}. It comprises $2,533$ images of $10$ classes shared by four different domains: Amazon (\textbf{A}), Caltech (\textbf{C}), DSLR (\textbf{D}), and Webcam (\textbf{W}). We follow~\cite{msda} to use one domain as the target domain and the others as the source domains. We evaluate on $4$ MSDA tasks. 
We follow the recent work~\cite{msda} to report the mean classification result of the unlabeled target domain data over five random trials and use the ImageNet~\cite{imagenet} pre-trained ResNet-101~\cite{resnet} as the base network. Other settings and implementation details are the same as those used in the single-source setting (cf. Section \ref{SecSetups}). 

\begin{table}[!t]
	\begin{center}
		\caption{Results (\%) on Office-Caltech10 for multi-source domain adaptation based on ResNet-101.
		}
		\label{table:results_office_caltech10} 
		\resizebox{0.8\linewidth}{!}{
			\begin{tabular}{lccccc}
				\hline
				Methods & A,C,D $\rightarrow$ W & A,C,W $\rightarrow$ D & A,D,W $\rightarrow$ C & C,D,W $\rightarrow$ A & Avg \\
				\hline
				Source Only~\cite{resnet}                 & 99.1 & 98.2 & 85.4 & 88.7 & 92.9 \\
				
				DAN~\cite{dan}               & 99.3 & 98.2 & 89.7 & 94.8 & 95.5 \\
				
				%DCTN~\cite{dctn}             & 99.4 & 99.0 & 90.2 & 92.7 & 95.3 \\
				
				JAN~\cite{jan}               & 99.4 & 99.4 & 91.2 & 91.8 & 95.5 \\
				
				DANN~\cite{dann}             & 99.3 & 98.7 & 90.7 & 91.1 & 95.0 \\
				
				MCD~\cite{mcd}               & 99.5 & 99.1 & 91.5 & 92.1 & 95.6 \\
				
				MEDA~\cite{meda}             & 99.3 & 99.2 & 91.4 & 92.9 & 95.7 \\
				
				M$^3$SDA~\cite{msda}         & 99.5 & 99.2 & 92.2 & 94.5 & 96.4 \\
				
				\hline
				\textbf{DisClusterDA}        & \textbf{100.0} & \textbf{100.0} & \textbf{96.1} & \textbf{96.4} & \textbf{98.1} \\
				\hline
			\end{tabular}
		}
	\end{center} 
\end{table}

Results on Office-Caltech10 are reported in Table \ref{table:results_office_caltech10}, where results of existing methods are quoted from their respective papers or~\cite{msda}. We can observe that DisClusterDA significantly outperforms all compared methods on all transfer tasks. The state-of-the-art MSDA method~\cite{msda} explicitly aligns each of the source domains not only with the target domain but also with each other and adopts ensemble schemas in the testing phase. In contrast, %we do not use the source domain labels since our proposed DisClusterDA can implicitly achieve feature alignment between any two domains. 
%In multi-source domain adaptation, 
our DisClusterDA works by implicitly achieving feature alignment between any two domains, although we do not use the source domain labels. More specifically, a common set of class centroids is considered in our source Fisher loss ${\cal{L}}_{Fisher}^s(F)$, which is consistent with the fact that all source domains share the label space. By minimizing ${\cal{L}}_{Fisher}^s(F)$, the samples of the same class from different source domains are pulled close to the same centroid, thus implicitly fulfilling the feature alignment between different source domains. On the other hand, the source and target ordering losses ${\cal{L}}_{ordering}^s$ and ${\cal{L}}_{ordering}^t$ are computed on cluster centroids from the respective domains. By minimizing them, the corresponding source and target centroids are classified as the same class, thus implicitly achieving the feature alignment between the source and target domains. 
%In summary, the impressive results in Table \ref{table:results_office_caltech10} can be attributed to our implicit manner of joint classification and clustering training. 
Empirical results verify the validity of our method with no damage of intrinsic discriminative structures in dealing with multi-source scenarios.

\section{Conclusion and Future Work}
\label{SecConclusion}

In this work, we pursue an alternative direction to learn classification of target data directly, with no explicit domain-level or class-level feature alignments. To this end, we propose a novel method of distilled discriminative clustering for unsupervised domain adaptation, termed DisClusterDA. It uses clustering objectives based on the proposed adaptive filtering entropy minimization loss, a soft Fisher-like criterion, and additionally the centroid classification via cluster ordering, thus effectively learning target-specific discriminative features. To regularize discriminative clustering of target data, we jointly train the network using parallel, supervised learning objectives over the labeled source data, which are mainly used as structural constraints. %Given shared learning of the network, our method can thus be viewed as learning to align the two domains implicitly, in contrast to most of the existing methods that strive to perform explicit feature alignment. 
Empirically, our method achieves the new state of the art on five popular benchmark datasets, verifying the efficacy of DisClusterDA%; it is also interesting to observe that in our DisClusterDA framework, adding an additional loss term of explicit class-level feature alignment produces degraded performance, corroborating our motivation in this work; certainly, more careful studies in different algorithmic frameworks are to be conducted
. 

DisClusterDA cannot be directly applied to the partial and open-set domain adaptation settings since the two domains have different label spaces and the underlying assumption behind our method is that the number of source classes is equal to that of target clusters. Future research is desired to address this issue. 
It is also desired to explore knowledge transfer without damage to the discriminative structures of target data. This may require a principled metric to measure how much a domain adaptation method hurts the intrinsic target structures, which is expected to be small. %, and seeking knowledge transfer methods with a low value of such a metric.

\section*{Acknowledgment}

This work was supported in part by the National Natural Science Foundation of China (Grant No.: 61771201), the Program for Guangdong Introducing Innovative and Enterpreneurial Teams (Grant No.: 2017ZT07X183), and the Guangdong R\&D key project of China (Grant No.: 2019B010155001).

%\section*{References}

\begin{small}
   	\bibliographystyle{elsarticle-num}
   	\bibliography{references}
\end{small}

\noindent\textbf{Hui Tang} received the B.E. degree in School of Electronic and Information Engineering from South China University of Technology, China, in 2018. She is currently pursuing the Ph.D. degree in School of Electronic and Information Engineering from South China University of Technology. Her research interests are in computer vision and pattern recognition.

\noindent\textbf{Yaowei Wang} received the Ph.D. degree in Computer Science from the Graduate University of Chinese Academy of Sciences in 2005. He worked at the Department of Electronics Engineering, Beijing Institute of Technology from 2005 to 2019. From 2014 to 2015, he worked as an academic Visitor at the vision lab of Queen Mary University of London. He was a professor at National Engineering Laboratory for Video Technology Shenzhen (NELVT), Peking University Shenzhen Graduate School in 2019. He is currently an associate professor with the Peng Cheng Laboratory, Shenzhen, China. 
His research interests include machine learning and multimedia content analysis and understanding. 
%He was the recipient of the second prize of the National Technology Invention in 2017 and the first prize of the CIE Technology Invention in 2015. His team was ranked as one of the best performers in the TRECVID CCD/SED tasks from 2009 to 2012 and in PETS 2012.

\noindent\textbf{Kui Jia} received the B.E. degree from Northwestern Polytechnic University, Xi’an, China, in 2001, the M.E. degree from the National University of Singapore, Singapore, in 2004, and the Ph.D. degree in computer science from the Queen Mary University of London, London, U.K., in 2007.
He was with the Shenzhen Institute of Advanced Technology of the Chinese Academy of Sciences, Shenzhen, China, Chinese University of Hong Kong, Hong Kong, the Institute of Advanced Studies, University of Illinois at Urbana-Champaign, Champaign, IL, USA, and the University of Macau, Macau, China. He is currently a Professor with the School of Electronic and Information Engineering, South China University of Technology, Guangzhou, China. His recent research focuses on theoretical deep learning and its applications in vision and robotic problems, including deep learning of 3D data and deep transfer learning.
	
\end{document}